\documentclass[10pt,twocolumn,letterpaper]{article}
\usepackage[pagenumbers]{cvpr}
\usepackage[T1]{fontenc}

\ifdefined\XeTeXversion
  \catcode`\–=\active \def–{\textendash}
  \catcode`\’=\active \def’{\textquoteright}
\fi

\usepackage{microtype}

\renewcommand{\paragraph}[1]{\vspace{.4em}\noindent\textbf{#1}}

\usepackage{booktabs}
\usepackage{multirow}
\usepackage{pgfplots}
\pgfplotsset{compat=1.18}

\usetikzlibrary{patterns}
\definecolor{segmaskblue}{HTML}{0079A8}
\definecolor{segmaskyellow}{HTML}{D6AA00}
\definecolor{segmaskgray}{HTML}{666666}
\definecolor{segmaskmagenta}{HTML}{CC79A7}
\DeclareRobustCommand{\segmasktp}{\tikz[baseline=-0.4ex]{\fill[segmaskblue] (0,0) rectangle (0.95em,0.65em);}}
\DeclareRobustCommand{\segmaskgt}{\tikz[baseline=-0.4ex]{\draw[segmaskyellow,line width=0.5pt,dash pattern=on 1.3pt off 0.8pt] (0,0) rectangle (0.95em,0.65em);}}
\definecolor{segmaskorange}{HTML}{E69F00}\DeclareRobustCommand{\segmaskhardfp}{\tikz[baseline=-0.4ex]{\fill[segmaskorange] (0,0) rectangle (0.95em,0.65em);\clip (0,0) rectangle (0.95em,0.65em);\draw[overlay,white,line width=0.3pt] (-0.15em,0)--(0.5em,0.65em) (0.45em,0)--(1.1em,0.65em);}}
\DeclareRobustCommand{\segmaskotherfp}{\tikz[baseline=-0.4ex]{\fill[segmaskgray] (0,0) rectangle (0.95em,0.65em);\draw[white,line width=0.18pt] (0.175em,0.325em) circle[radius=0.035em] (0.475em,0.325em) circle[radius=0.035em] (0.775em,0.325em) circle[radius=0.035em];}}
\DeclareRobustCommand{\segmaskfn}{\tikz[baseline=-0.4ex]{\fill[segmaskmagenta] (0,0) rectangle (0.95em,0.65em);\clip (0,0) rectangle (0.95em,0.65em);\draw[overlay,white,line width=0.3pt] (-0.15em,0.65em)--(0.5em,0) (0.45em,0.65em)--(1.1em,0);}}

\definecolor{cvprblue}{rgb}{0.21,0.49,0.74}
\usepackage[breaklinks,colorlinks,allcolors=cvprblue]{hyperref}
\hypersetup{
  pdftitle={SegPFN: Training-Free In-Context Segmentation with Tabular Prior-Data Fitted Networks},
  pdfauthor={Jiajian Ma, Yihao Qiao, Ke Chen, Kaiyue Feng, Wushuang Rui, Abhishek Jha, Aarti Darji, Yiqiu Shen}
}

\title{SegPFN: Training-Free In-Context Segmentation\\with Tabular Prior-Data Fitted Networks}
\author{%
  Jiajian Ma\textsuperscript{1}\quad Yihao Qiao\textsuperscript{1}\quad
  Ke Chen\textsuperscript{1}\quad Kaiyue Feng\textsuperscript{1}\\
  Wushuang Rui\textsuperscript{1}\quad Abhishek Jha\textsuperscript{1}\quad
  Aarti Darji\textsuperscript{1}\quad Yiqiu Shen\textsuperscript{2,1}\\[2pt]
  {\normalsize\textsuperscript{1}New York University\qquad
  \textsuperscript{2}NYU Langone Health}\\[1pt]
  {\normalsize \href{https://github.com/nyu-shenlab/SegPFN}{\textsf{github.com/nyu-shenlab/SegPFN}}}%
}

\makeatletter
\let\segpfn@originalmaketitle\@maketitle
\patchcmd{\segpfn@originalmaketitle}{\vskip .375in}{\kern -15pt}{}{}
\patchcmd{\segpfn@originalmaketitle}{\vspace*{24pt}}{\vspace*{7pt}}{}{}
\patchcmd{\abstract}{\vspace*{12pt}}{\vspace*{6pt}}{}{}
\patchcmd{\endabstract}{\vspace*{12pt}}{}{}{}
\renewcommand{\@maketitle}{%
  \segpfn@originalmaketitle
  \vspace{-27pt}%
  \noindent\begin{minipage}{\textwidth}
    \centering
    \captionsetup{hypcap=false}
    \includegraphics[width=\linewidth]{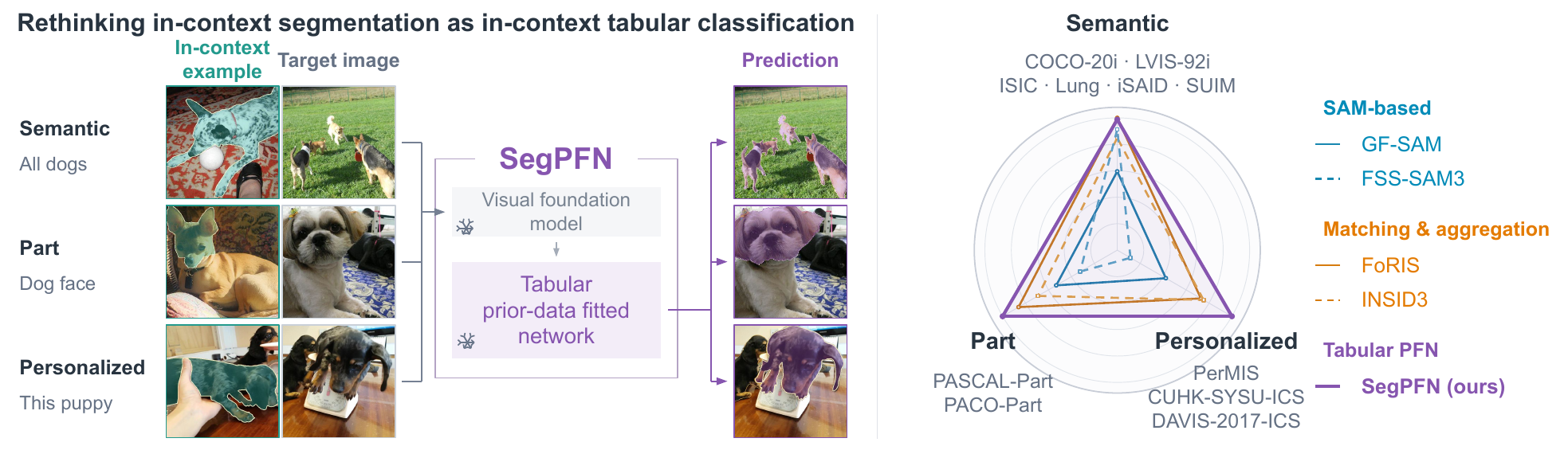}
    \captionof{figure}{\textbf{SegPFN reformulates in-context segmentation as tabular classification.} Left: Annotated reference images specify a semantic category, an object part, or a specific instance. Using patch features extracted by a frozen visual foundation model, a frozen tabular prior-data fitted network (PFN) predicts the target mask conditioned on the reference annotations. Right: Comparison of training-free methods across the three segmentation tasks, with evaluation datasets listed along each axis.
    }
    \label{fig:teaser}
  \end{minipage}
  \par\vspace{4pt}%
}
\makeatother

\begin{document}
\maketitle
\makeatletter
\let\segpfn@savedtextbottom\@textbottom
\let\segpfn@savedtexttop\@texttop
\raggedbottom
\AddToHookNext{shipout/after}{%
  \global\let\@textbottom\segpfn@savedtextbottom
  \global\let\@texttop\segpfn@savedtexttop
}
\makeatother
\begin{abstract}
\hyphenpenalty=140\relax
In-context segmentation (ICS) uses one or a few annotated reference images to specify what to segment in a target image. Existing training-free ICS approaches typically rely on manually designed inference procedures, which may limit how flexibly reference information is combined. Learned in-context inference offers an adaptive alternative, but acquiring this capability through segmentation-specific training can compromise generalization to unseen domains. We therefore ask: can a pretrained in-context learner infer what to segment from annotated references without segmentation-specific training? We introduce SegPFN, a training-free framework that reformulates ICS as tabular classification over frozen visual features. A frozen tabular prior-data fitted network (PFN) uses labeled reference patches as context to classify patches in the target image. We comprehensively benchmark SegPFN against major existing ICS methods across 11 datasets spanning semantic, part, and personalized segmentation. SegPFN improves overall mean intersection-over-union (mIoU) by 4.7 points over the strongest overall baseline and achieves the highest average mIoU in four of five task–shot settings. Further analyses show that hard-negative reference patches improve fine-grained target discrimination, while joint conditioning enables SegPFN to use multiple references effectively.
These results show that SegPFN offers a simple and effective paradigm for \mbox{training-free in-context segmentation}.

\end{abstract}

\section{Introduction}
\label{sec:intro}

Image segmentation identifies pixels belonging to regions of interest. Conventional supervised approaches typically learn a predefined segmentation task from annotated training data~\citep{long2015fully}. \textit{In-context segmentation (ICS)} offers a more flexible alternative: given one or a few annotated reference images, a model identifies corresponding regions in an unlabeled target image \textit{without task-specific parameter updates at inference time}~\citep{wang2023seggpt,liu2024matcher}. ICS supports tasks at different levels of specificity: \textit{semantic segmentation} identifies all objects of a specified category (e.g., all dogs), \textit{part segmentation} identifies specific object parts (e.g., dog ears), and \textit{personalized segmentation} identifies a particular object instance (e.g., one specific dog). This flexibility has enabled broad applications across medical imaging~\citep{xie2025eicseg}, video object segmentation~\citep{wang2023seggpt}, and personalized image generation~\citep{zhang2024persam}.


Early approaches learn to perform ICS through
segmentation-specific
training~\citep{wang2023seggpt,liu2024sine,meng2024segic}.
Recent advances in visual foundation models, such as
DINOv3~\citep{simeoni2025dinov3} and the Segment Anything Model
(SAM)~\citep{kirillov2023sam}, have enabled ICS without additional segmentation-specific training. These training-free approaches broadly follow two strategies. 
The first uses visual feature matching between reference and
target images to generate prompts for pretrained segmentation
models, such as SAM, or to select relevant candidate 
masks~\citep{zhang2024persam,liu2024matcher,zhang2024gfsam}. The second predicts target masks directly from pretrained visual
features through predefined matching, aggregation, or statistical
estimation procedures~\citep{cuttano2026insid3,hu2026foris,park2026frost}. 
These training-free methods typically rely on manually designed inference procedures, which may limit how flexibly information from multiple references is combined.
Learned in-context inference offers an adaptive alternative, but acquiring this capability through segmentation-specific training can compromise generalization to unseen domains~\citep{cuttano2026insid3}.
This raises a natural question: \textbf{\textit{Can a pretrained in-context learner infer what to segment from annotated references without segmentation-specific training?}}

Tabular prior-data fitted networks (PFNs) offer a promising solution. They are pretrained on diverse synthetic tabular tasks to perform in-context learning: given labeled examples as context, they predict labels for new samples without updating model parameters at inference time~\citep{muller2022pfn,hollmann2025tabpfn}. Although pretrained only on tabular data, these models have also been shown to perform classification using global features extracted from other modalities, including images, text, and audio~\citep{bergstrom2026comet,zhang2026tabpfntabulardatacalibration}. These findings motivate extending tabular in-context learning from sample-level classification to dense prediction over image patches.

We propose SegPFN (In-Context \textbf{Seg}mentation with Tabular
\textbf{P}rior-Data \textbf{F}itted \textbf{N}etworks), a
training-free framework that reformulates ICS as in-context
tabular classification. A frozen visual encoder extracts
patch-level features from reference and target images.
Each patch becomes a sample (row), its feature dimensions
serve as predictors (columns), and reference masks provide
foreground or background labels for the reference patches.
A pretrained tabular PFN then uses sampled, labeled reference
patches as context to predict foreground probabilities for
target patches without updating its parameters. The PFN
jointly conditions on positive and negative patches across
references within a shared context, allowing the annotations
to specify both what to segment and what to exclude.
Finally, we apply feature-based and spatial refinement to improve prediction consistency and segmentation boundaries.

\noindent Our contributions are threefold:
\textbf{(1)} We introduce SegPFN, a training-free framework that reformulates ICS as a tabular classification problem and transfers the in-context prediction capability of a tabular PFN to dense visual prediction.
\textbf{(2)} We benchmark SegPFN against major
existing ICS methods across 11 datasets spanning semantic, part,
and personalized segmentation. SegPFN improves overall mean intersection-over-union (mIoU) by 4.7 points over the strongest overall baseline and achieves the highest average mIoU in four of five evaluated task–shot settings, with particular strengths in fine-grained target discrimination and effective use of multiple references. 
\textbf{(3)} We introduce two complementary personalized segmentation benchmarks: DAVIS-2017-ICS for selecting different instances within fixed image pairs by changing only the reference mask, and CUHK-SYSU-ICS for segmenting the same person across views.
We will release the code and task definitions for both benchmarks.

\section{Related Work}
\label{sec:related_work}

\paragraph{In-Context Segmentation.}
In-context segmentation (ICS) uses annotated reference
images to specify semantic categories, object parts, or
individual instances to segment in a target image, without task-specific parameter updates at inference time. 
Several methods learn this capability through segmentation-specific training.
Painter~\cite{wang2023painter} and SegGPT~\cite{wang2023seggpt} train general-purpose models for visual prediction from examples, while SegIC~\cite{meng2024segic} and SINE~\cite{liu2024sine} learn reference-guided segmentation.
FSSAM~\cite{xu2025unlocking} and SANSA~\cite{cuttano2025sansa} require adaptation of SAM~2.

Recent work has increasingly focused on training-free
ICS, which requires no additional segmentation-specific training.
One line builds on pretrained prompt-based segmentation models.
PerSAM~\cite{zhang2024persam}, Matcher~\cite{liu2024matcher}, and GF-SAM~\cite{zhang2024gfsam} use reference information to guide mask generation with SAM~\cite{kirillov2023sam} or to select and aggregate its candidate masks.
FSS-SAM3~\cite{tsai2026fewshotsemanticsegmentationmeets} instead places reference and target images on the same canvas and predicts masks with SAM~3.

Another line of training-free methods predicts masks directly from frozen visual features, without relying on pretrained segmentation decoders.
INSID3~\cite{cuttano2026insid3} combines feature matching with region grouping, while FoRIS~\cite{hu2026foris} progressively refines foreground predictions through matching and aggregation.
FSSDINO~\cite{zakir2026revealingsemanticselectiongap} uses feature prototypes and refinement, and FROST~\cite{park2026frost} estimates nonparametric foreground-to-background density ratios.
These methods enable ICS by combining pretrained visual features with manually designed inference procedures.
SegPFN instead transfers the in-context prediction capability of a tabular PFN pretrained on synthetic tabular tasks to visual segmentation.


\paragraph{Tabular Prior-Data Fitted Networks.}
Prior-data fitted networks (PFNs) are neural networks trained on synthetic tasks sampled from a prior distribution to learn in-context prediction~\cite{muller2022pfn}. Given labeled examples as context, they approximate Bayesian posterior predictive distributions for new inputs without updating their parameters. Tabular PFNs specialize this framework for tabular classification and regression by pretraining on synthetic tabular datasets. The TabPFN family~\cite{hollmann2023tabpfn,hollmann2025tabpfn,jager2026tabpfn35} is a prominent example, outperforming extensively tuned machine learning baselines on many small-data benchmarks.

Recent studies show that tabular PFNs can also operate on global features extracted from other modalities by pretrained encoders.
CoMET~\cite{bergstrom2026comet} combines frozen modality encoders, principal component analysis (PCA), and tabular foundation models for training-free multimodal classification.
Zhang et al.~\cite{zhang2026tabpfntabulardatacalibration} evaluate TabPFN as a classification head for frozen image, text, and audio features.
Both studies focus on sample-level classification.

Beyond sample-level classification, tabular PFNs have been applied to pixel-level prediction in remote sensing. 
Ta et al.~\cite{ta2026sarflood} use handcrafted statistics from synthetic aperture radar time series as pixel features and a small set of labeled locations as context for flood mapping. 
WSPE~\cite{zhao2026wspe} uses TabPFN to classify pixels based on spectral and texture features. These predictions guide the selection of point prompts for
SAM~2~\cite{ravi2025sam2}, which generates pseudo-labels
for training a separate segmentation model.
In contrast, SegPFN applies a tabular PFN to in-context segmentation across semantic, part, and personalized tasks, with dense visual encoder features as input and annotated reference images specifying what to segment.

\section{In-Context Segmentation with SegPFN}
\label{sec:method}

\paragraph{Task definition.}
Given $K$ annotated reference images
$\mathcal{R}=\{(I_k^r,M_k^r)\}_{k=1}^{K}$ and a target
image $I^t$, our goal is to predict the target mask
${M}^t$.
Here, $I_k^r,I^t\in\mathbb{R}^{H\times W\times3}$ and
$M_k^r,{M}^t\in\{0,1\}^{H\times W}$, with $1$ and $0$
denoting foreground and background, respectively.

\begin{figure*}[t]
    \centering
    \includegraphics[
        width=\textwidth,
        clip
    ]{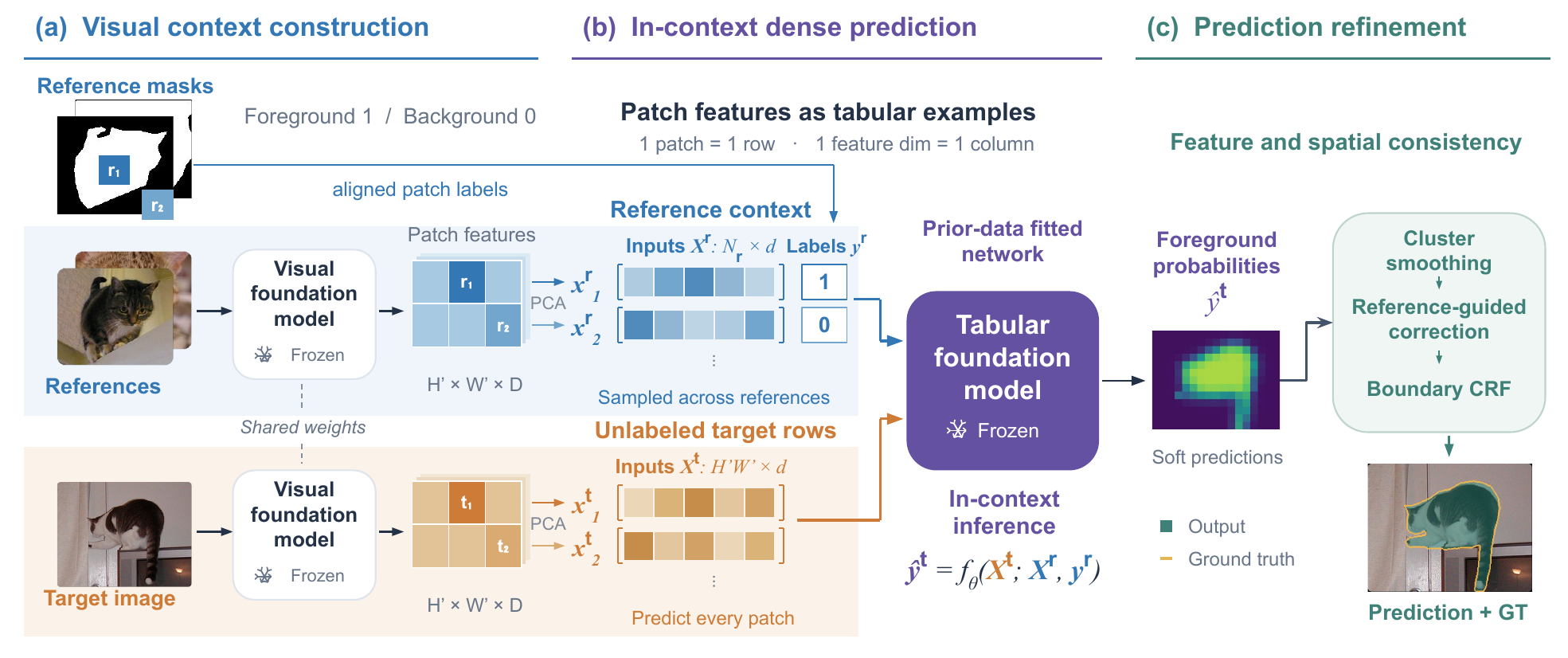}
    \caption{
        \textbf{Overview of SegPFN.}
        (a) Reference and target patches are represented by compact
        visual features. Sampled reference patches and their
        mask-derived labels form the context.
        (b) A frozen tabular PFN predicts foreground probabilities for target patches,
        treating each patch as a row and each feature dimension
        as a column.
        (c) Cluster smoothing, reference-guided correction, and a
        boundary conditional random field (CRF) produce the final mask.
        Ground truth (GT) is shown only for comparison.
    }
    \label{fig:segpfn-overview}
\end{figure*}

SegPFN formulates ICS as dense in-context tabular classification
over frozen visual features.
As illustrated in Fig.~\ref{fig:segpfn-overview}, we
construct a labeled reference context
(Sec.~\ref{sec:features}), predict target foreground
probabilities with a pretrained tabular PFN
(Sec.~\ref{sec:pfn_prediction}), and refine the predictions
using feature and spatial consistency
(Sec.~\ref{sec:refinement}).
Both the visual encoder and the tabular PFN remain frozen.

\subsection{Visual Context Construction}
\label{sec:features}

\paragraph{Patch representation.}
A frozen visual encoder maps each reference image $I_k^r$
and the target image $I^t$ to an $H'\times W'$ grid of
$D$-dimensional patch features.
After patch-wise $\ell_2$ normalization, we apply principal
component analysis (PCA) to reduce the feature dimension
from $D$ to $d$, accommodating the tabular PFN's input
feature budget.
PCA is fitted on all valid patches from the original
reference images before context sampling.
The same projection is applied to both reference and target features, including those from augmented views.

\paragraph{Reference context sampling.}
We align the reference masks $M_k^r$ to the encoder grid
to obtain binary patch labels.
To bound the PFN context length and inference cost, we
sample at most $B$ labeled reference patches.
We allocate the sampling budget in proportion to the
foreground and background counts across the original
references, enforcing a minimum quota per class when
sufficient patches are available.
Within each class, we distribute the quota equally across
reference views and sample patches uniformly without
replacement.

The sampled reference features and their labels form the
context
$\mathbf{X}^r\in\mathbb{R}^{N_r\times d}$ and
$\mathbf{y}^r\in\{0,1\}^{N_r}$, where $N_r\leq B$. For $K>1$, patches from all references form a shared context, enabling joint reference conditioning under the same total budget $B$.

\subsection{In-Context Dense Prediction}
\label{sec:pfn_prediction}

We treat each patch as a tabular example, with its
$d$ projected feature dimensions forming the columns.
We flatten the projected feature map of $I^t$ into
$\mathbf{X}^t\in\mathbb{R}^{H'W'\times d}$.
Conditioned on the reference context
$(\mathbf{X}^r,\mathbf{y}^r)$, a pretrained tabular PFN
predicts foreground probabilities for all target patches:
\begin{equation}
    \hat{\mathbf{y}}^t
    =
    f_{\theta}(\mathbf{X}^t;\mathbf{X}^r,\mathbf{y}^r)
    \in[0,1]^{H'W'},
    \label{eq:segpfn_formulation}
\end{equation}
where $\theta$ denotes the frozen PFN parameters and
$\hat{y}_i^t$ is the foreground probability of target
patch $i$.
We reshape $\hat{\mathbf{y}}^t$ to the $H'\times W'$ patch
grid for refinement.
\subsection{Prediction Refinement}
\label{sec:refinement}

To improve prediction consistency across visually similar target patches, we propose \emph{cluster smoothing} for PFN predictions. We obtain patch groups through agglomerative clustering of the original $D$-dimensional, $\ell_2$-normalized features. Let $G(i)$ denote the cluster containing target patch $i$. We then average its predicted foreground probability $\hat{y}_i^t$ with the cluster mean:
\begin{equation}
    s_i
    =
    \frac{1}{2}
    \underbrace{\hat{y}_i^t}_{\text{Original score}}
    +
    \frac{1}{2}
    \underbrace{
        \left(
            \frac{1}{|G(i)|}
            \sum_{j\in G(i)} \hat{y}_j^t
        \right)
    }_{\text{Cluster mean}}.
    \label{eq:feature_consistency}
\end{equation}

We min--max normalize the smoothed scores $s_i$ and apply
a \emph{reference-guided correction} based on
FoRIS~\cite{hu2026foris}. 
We then upsample and threshold the corrected score map
to obtain a binary mask, which is refined with the conditional random field (CRF)~\cite{krahenbuhl2011efficient} to produce the final prediction $\hat{M}^t$.
Details are provided in the supplementary
material.

\section{Experiments}
\label{sec:experiments}

We investigate three research questions. First, can a pretrained tabular PFN perform ICS without segmentation-specific training? We evaluate SegPFN's generalization across semantic, part, and personalized segmentation in diverse image domains (Sec.~\ref{sec:generalization}).
Second, what are SegPFN's advantages and what drives them? We examine its ability to distinguish the requested instance from same-category non-target instances (\emph{distractors}) and investigate the roles of hard-negative reference patches and joint reference conditioning (Sec.~\ref{sec:advantages}). Third, which components contribute to its performance? We compare PFN-based prediction with conventional classifiers and evaluate the effects of prediction refinement (Sec.~\ref{sec:ablations}).

\paragraph{Datasets.}
We evaluate SegPFN on 11 datasets spanning semantic, part,
and personalized segmentation.
For semantic segmentation, we use COCO-$20^i$~\cite{lin2014coco,nguyen2019feature}, LVIS-$92^i$~\cite{gupta2019lvis,liu2024matcher},
ISIC~\cite{codella2019skin,tschandl2018ham10000}, Lung~\cite{jaeger2014tuberculosis,candemir2014lung}, iSAID~\cite{zamir2019isaid}, and SUIM~\cite{islam2020suim}, covering natural, medical,
aerial, and underwater images.
For part segmentation, we use PASCAL-Part~\cite{chen2014detect} and PACO-Part~\cite{ramanathan2023paco}.
For personalized segmentation, we use PerMIS~\cite{samuel2024waldo} and two
additional benchmarks that we construct:
DAVIS-2017-ICS and CUHK-SYSU-ICS.

\paragraph{Additional Personalized Segmentation Benchmarks.}
Personalized segmentation requires identifying a specific object instance while excluding other similar instances of the same category. PerMIS~\cite{samuel2024waldo} evaluates this ability using reference and target images from the same video. To further test target discrimination, we introduce two complementary benchmarks. DAVIS-2017-ICS evaluates whether a model can select different target instances from the same image pair based solely on changes to the reference mask. CUHK-SYSU-ICS evaluates whether a model can identify the same person across different views containing other people.

We construct the \textbf{DAVIS-2017-ICS} benchmark using images, instance masks, and category mappings from the DAVIS 2017 training and validation splits~\cite{ponttuset2017davis}. We select temporally separated reference--target frame pairs in which the target frame contains at least two instances of the same category. Each eligible instance appearing in both frames defines a separate task, with its reference mask specifying the target and its target-frame mask providing ground truth. The image pair remains fixed across these tasks. Only the requested instance changes. We manually verify instance correspondence using the provided instance IDs. The resulting benchmark contains 53 tasks from 20 videos across nine categories.

We construct the \textbf{CUHK-SYSU-ICS} benchmark by combining person identity annotations from the CUHK-SYSU training split~\cite{xiao2017personsearch} with manually annotated polygon masks from maskPS~\cite{zheng2020segmentation}. We retain identities with valid masks in at least two images and use bounding boxes to verify that each target image contains at least one other person. We manually verify identities, reference--target assignments, and image quality, prioritizing pairs from different scenes. The benchmark contains 151 identity-segmentation tasks from 113 reference--target pairs.

Both benchmarks retain full, uncropped images. All methods receive identical reference--target pairs and reference masks. Target-image masks are used only for benchmark construction, verification, and evaluation, and are not provided to the methods. Additional construction details are provided in the supplementary material.

\paragraph{Baselines.}
We compare SegPFN with three groups of methods:
(1) ICS methods with additional segmentation-specific training, including
SINE~\cite{liu2024sine},
SegIC~\cite{meng2024segic}, and
FSSAM~\cite{xu2025unlocking};
(2) training-free methods built on pretrained segmentation
models, including Matcher~\cite{liu2024matcher},
GF-SAM~\cite{zhang2024gfsam}, and
FSS-SAM3~\cite{tsai2026fewshotsemanticsegmentationmeets};
and (3) training-free methods operating on frozen visual
features, including
FSSDINO~\cite{zakir2026revealingsemanticselectiongap},
INSID3~\cite{cuttano2026insid3},
FROST~\cite{park2026frost}, and
FoRIS~\cite{hu2026foris}.

\begingroup\captionsetup{hypcap=false}\begin{table*}[!t]
\centering
\caption{One-shot and five-shot semantic segmentation (mIoU in \%, $\uparrow$). Avg.~(6) equally weights the six datasets; Total Avg. is the mean of the one-shot and five-shot averages. Bold/underline mark the best/second-best eligible scores. Gray indicates overlap between training and evaluation dataset sources; \S\ marks published Matcher results from INSID3. These entries and affected averages are excluded from ranking.}
\label{tab:semantic-results}
\footnotesize
\setlength{\tabcolsep}{1.5pt}
\renewcommand{\arraystretch}{1.15}
\begin{tabular*}{\textwidth}{@{\extracolsep{\fill}}l*{15}{r}@{}}
\toprule
\multirow{2}{*}{Method} & \multicolumn{2}{c}{COCO-$20^i$} & \multicolumn{2}{c}{LVIS-$92^i$} & \multicolumn{2}{c}{ISIC} & \multicolumn{2}{c}{Lung} & \multicolumn{2}{c}{iSAID} & \multicolumn{2}{c}{SUIM} & \multicolumn{2}{c}{Avg.~(6)} & \multirow{2}{*}{\shortstack{Total\\Avg.}} \\
\cmidrule(lr){2-3}\cmidrule(lr){4-5}\cmidrule(lr){6-7}\cmidrule(lr){8-9}\cmidrule(lr){10-11}\cmidrule(lr){12-13}\cmidrule(lr){14-15}
& 1-shot & 5-shot & 1-shot & 5-shot & 1-shot & 5-shot & 1-shot & 5-shot & 1-shot & 5-shot & 1-shot & 5-shot & 1-shot & 5-shot & \\
\midrule
\multicolumn{16}{@{}l}{\textit{Additional segmentation-specific training}} \\
SINE~\cite{liu2024sine} & \textcolor{gray}{64.53} & \textcolor{gray}{66.14} & 31.18 & 35.54 & 25.90 & 27.08 & 39.84 & 40.61 & 38.27 & 38.53 & 58.70 & 62.38 & \textcolor{gray}{43.07} & \textcolor{gray}{45.05} & \textcolor{gray}{44.06} \\
SegIC~\cite{meng2024segic} & \textcolor{gray}{74.05} & \textemdash & \textcolor{gray}{44.44} & \textemdash & 26.92 & \textemdash & 28.76 & \textemdash & 25.63 & \textemdash & \textbf{63.35} & \textemdash & \textcolor{gray}{43.86} & \textemdash & \textemdash \\
FSSAM~\cite{xu2025unlocking} & \textcolor{gray}{64.22} & \textcolor{gray}{72.14} & 37.27 & 49.71 & 53.50 & \underline{68.24} & 81.11 & 86.45 & 50.06 & 56.21 & 56.89 & 61.85 & \textcolor{gray}{57.18} & \textcolor{gray}{65.77} & \textcolor{gray}{61.47} \\
\addlinespace[3pt]
\multicolumn{16}{@{}l}{\textit{Training-free: mask-supervised visual pretraining}} \\
Matcher~\cite{liu2024matcher} & 52.80 & 60.70\textsuperscript{\S} & 32.97 & 40.00\textsuperscript{\S} & 37.35 & 35.00\textsuperscript{\S} & 71.39 & 71.20\textsuperscript{\S} & 35.07 & 34.30\textsuperscript{\S} & 44.10 & 50.60\textsuperscript{\S} & 45.61 & 48.63\textsuperscript{\S} & 47.12\textsuperscript{\S} \\
GF-SAM~\cite{zhang2024gfsam} & 58.80 & 67.16 & 35.20 & 44.26 & 49.19 & 56.12 & 51.23 & 49.90 & 48.45 & 54.79 & 54.93 & 60.52 & 49.63 & 55.46 & 52.55 \\
FSS-SAM3~\cite{tsai2026fewshotsemanticsegmentationmeets} & \textbf{67.07} & \textbf{71.81} & 31.23 & 47.12 & \textbf{70.27} & \textbf{71.79} & \textbf{87.90} & 71.96 & 30.21 & 45.21 & 39.26 & 56.88 & 54.32 & 60.80 & 57.56 \\
\addlinespace[3pt]
\multicolumn{16}{@{}l}{\textit{Training-free: self-supervised visual features}} \\
FSSDINO~\cite{zakir2026revealingsemanticselectiongap} & 46.99 & 58.54 & 12.07 & 16.37 & 55.97 & 62.92 & 53.02 & 53.20 & 43.40 & 53.26 & 50.40 & 28.13 & 43.64 & 45.40 & 44.52 \\
INSID3~\cite{cuttano2026insid3} & 56.67 & 65.49 & 41.89 & 47.22 & 56.20 & 64.24 & 78.88 & 79.37 & 51.62 & 58.59 & 54.80 & 63.40 & 56.68 & 63.05 & 59.86 \\
FROST~\cite{park2026frost} & 47.80 & 62.17 & 32.20 & 47.73 & 56.74 & 65.86 & 85.67 & 86.22 & 50.49 & 61.05 & 58.43 & 66.86 & 55.22 & 64.98 & 60.10 \\
FoRIS~\cite{hu2026foris} & \underline{60.93} & 68.16 & \textbf{43.07} & \underline{50.52} & \underline{62.79} & 67.66 & \underline{87.72} & \textbf{88.10} & \textbf{53.57} & \underline{61.71} & 59.10 & \underline{66.90} & \textbf{61.20} & \underline{67.18} & \underline{64.19} \\
\addlinespace[3pt]
\multicolumn{16}{@{}l}{\textit{Ours: frozen visual encoder and pretrained tabular PFN}} \\
SegPFN & 58.35 & \underline{68.63} & \underline{42.57} & \textbf{56.67} & 60.46 & 67.93 & 86.53 & \underline{86.92} & \underline{53.47} & \textbf{65.48} & \underline{59.36} & \textbf{69.66} & \underline{60.12} & \textbf{69.22} & \textbf{64.67} \\
\bottomrule
\end{tabular*}
\end{table*}\endgroup

\paragraph{Evaluation Protocol and Metrics.}
We use mean intersection-over-union (mIoU, \%) as the
primary evaluation metric. For semantic and part
segmentation, we follow the dataset splits and evaluation
protocols of INSID3~\cite{cuttano2026insid3} and
FoRIS~\cite{hu2026foris}, reporting both one-shot and
five-shot results. For personalized segmentation, we
follow the one-shot protocol established by INSID3~\cite{cuttano2026insid3} on
PerMIS~\cite{samuel2024waldo} and apply it to DAVIS-2017-ICS and CUHK-SYSU-ICS.
Only the specified target instance is treated as foreground,
with all other pixels considered background.
We report macro-averaged performance separately for each segmentation task type, assigning equal weight to each dataset. Overall mIoU equally weights the three task means after averaging the one-shot and five-shot results within semantic and part segmentation. Further details on evaluation settings and metric
aggregation are provided in the supplementary material.

\paragraph{Implementation Details.}
We use a frozen DINOv3-L as the visual encoder
~\cite{simeoni2025dinov3} and TabPFN~3.5
~\cite{jager2026tabpfn35} as the tabular PFN.
All images are processed at a resolution of $1024\times1024$.
We apply patch-wise $\ell_2$ normalization and reduce the
feature dimension to 128 using PCA fitted on the
reference-image features.
Each PFN context contains at most $1{,}024$ labeled
reference patches.
Both models remain frozen during inference.
Further details on PFN ensembling, reference augmentation,
and evaluation-specific settings are provided in the
supplementary material.

\subsection{Main Results}
\label{sec:generalization}

\begingroup\captionsetup{hypcap=false}\begin{table*}[!t]
\centering
\normalsize
\begin{minipage}[t]{\columnwidth}
\centering
\caption{One-shot and five-shot part segmentation (mIoU in \%, $\uparrow$). Avg.~(2) equally weights both datasets; Total Avg. is the mean of the one-shot and five-shot averages. Bold/underline mark the best/second-best eligible scores. \S\ marks published Matcher results from INSID3 and affected averages, excluded from ranking.}
\label{tab:part-results}
\footnotesize
\setlength{\tabcolsep}{1.6pt}
\renewcommand{\arraystretch}{1.13}
\begin{tabular*}{\columnwidth}{@{\extracolsep{\fill}}l*{7}{r}@{}}
\toprule
\multirow{2}{*}{Method} & \multicolumn{2}{c}{PASCAL-Part} & \multicolumn{2}{c}{PACO-Part} & \multicolumn{2}{c}{Avg.~(2)} & \multirow{2}{*}{\shortstack{Total\\Avg.}} \\
\cmidrule(lr){2-3}\cmidrule(lr){4-5}\cmidrule(lr){6-7}
& 1-shot & 5-shot & 1-shot & 5-shot & 1-shot & 5-shot & \\
\midrule
\multicolumn{8}{@{}l}{\textit{Additional segmentation-specific training}} \\
SINE~\cite{liu2024sine} & 34.86 & 34.79 & 23.22 & 25.33 & 29.04 & 30.06 & 29.55 \\
SegIC~\cite{meng2024segic} & 40.07 & \textemdash & 24.97 & \textemdash & 32.52 & \textemdash & \textemdash \\
FSSAM~\cite{xu2025unlocking} & 46.32 & 61.10 & 30.99 & 40.35 & 38.66 & 50.72 & 44.69 \\
\addlinespace[3pt]
\multicolumn{8}{@{}l}{\textit{Training-free: mask-supervised pretraining}} \\
Matcher~\cite{liu2024matcher} & 42.89 & 45.80\textsuperscript{\S} & 34.72 & 33.60\textsuperscript{\S} & 38.81 & 39.70\textsuperscript{\S} & 39.26\textsuperscript{\S} \\
GF-SAM~\cite{zhang2024gfsam} & 44.69 & 51.81 & 36.73 & 41.91 & 40.71 & 46.86 & 43.79 \\
FSS-SAM3~\cite{tsai2026fewshotsemanticsegmentationmeets} & 36.03 & 56.22 & 20.67 & 38.37 & 28.35 & 47.30 & 37.83 \\
\addlinespace[3pt]
\multicolumn{8}{@{}l}{\textit{Training-free: self-supervised features}} \\
FSSDINO~\cite{zakir2026revealingsemanticselectiongap} & 30.19 & 32.41 & 19.03 & 22.07 & 24.61 & 27.24 & 25.93 \\
INSID3~\cite{cuttano2026insid3} & 49.62 & 58.20 & 38.47 & 47.02 & 44.04 & 52.61 & 48.33 \\
FROST~\cite{park2026frost} & \underline{55.85} & \textbf{67.53} & 36.99 & 50.90 & 46.42 & \underline{59.22} & 52.82 \\
FoRIS~\cite{hu2026foris} & 55.80 & 63.33 & \underline{42.53} & \underline{51.07} & \underline{49.17} & 57.20 & \underline{53.19} \\
\addlinespace[3pt]
\multicolumn{8}{@{}l}{\textit{Ours: frozen visual encoder and tabular PFN}} \\
SegPFN & \textbf{57.45} & \underline{67.09} & \textbf{46.50} & \textbf{57.55} & \textbf{51.97} & \textbf{62.32} & \textbf{57.15} \\
\bottomrule
\end{tabular*}
\end{minipage}\hfill
\begin{minipage}[t]{\columnwidth}
\centering
\caption{One-shot personalized segmentation (foreground mIoU in \%, $\uparrow$). Avg. equally weights the three datasets. Bold and underline indicate the best and second-best scores in each column.}
\label{tab:personalized}
\footnotesize
\setlength{\tabcolsep}{2.2pt}
\renewcommand{\arraystretch}{1.20}
\begin{tabular*}{\columnwidth}{@{\extracolsep{\fill}}lrrrr@{}}
\toprule
Method & PerMIS & \shortstack{CUHK-SYSU-\\ICS} & \shortstack{DAVIS-2017-\\ICS} & Avg. \\
\midrule
\multicolumn{5}{@{}l}{\textit{Additional segmentation-specific training}} \\
SINE~\cite{liu2024sine} & 42.48 & 26.20 & 27.69 & 32.12 \\
SegIC~\cite{meng2024segic} & 52.78 & 39.67 & 41.86 & 44.77 \\
FSSAM~\cite{xu2025unlocking} & 61.61 & 39.20 & 34.01 & 44.94 \\
\addlinespace[3pt]
\multicolumn{5}{@{}l}{\textit{Training-free: mask-supervised visual pretraining}} \\
Matcher~\cite{liu2024matcher} & 64.36 & \textbf{50.02} & 37.28 & 50.56 \\
GF-SAM~\cite{zhang2024gfsam} & 54.57 & 38.04 & 29.60 & 40.74 \\
FSS-SAM3~\cite{tsai2026fewshotsemanticsegmentationmeets} & 39.93 & 27.59 & 22.88 & 30.14 \\
\addlinespace[3pt]
\multicolumn{5}{@{}l}{\textit{Training-free: self-supervised visual features}} \\
FSSDINO~\cite{zakir2026revealingsemanticselectiongap} & 20.86 & 26.65 & 24.88 & 24.13 \\
INSID3~\cite{cuttano2026insid3} & \underline{67.17} & 42.91 & 38.75 & 49.61 \\
FROST~\cite{park2026frost} & 66.59 & 45.00 & \underline{50.18} & \underline{53.93} \\
FoRIS~\cite{hu2026foris} & 64.41 & 45.20 & 38.68 & 49.43 \\
\addlinespace[3pt]
\multicolumn{5}{@{}l}{\textit{Ours: frozen visual encoder and pretrained tabular PFN}} \\
SegPFN & \textbf{71.76} & \underline{49.71} & \textbf{55.92} & \textbf{59.13} \\
\bottomrule
\end{tabular*}
\par\vspace{4pt}
\raggedright\scriptsize Dataset construction and SegIC's dataset-specific inference settings are detailed in the supplementary material.
\end{minipage}
\end{table*}\endgroup

\paragraph{Semantic segmentation.} Table~\ref{tab:semantic-results} shows that SegPFN achieves the highest average across one- and five-shot settings (64.67), slightly exceeding FoRIS (64.19). In the one-shot setting, SegPFN scores 60.12, compared with 61.20 for FoRIS. However, in the five-shot setting, SegPFN achieves the highest average of 69.22, outperforming FoRIS by 2.04 points. It outperforms FoRIS on five of the six datasets, with the largest gains on LVIS (+6.15) and iSAID (+3.77), while FoRIS remains stronger on Lung.

\paragraph{Part segmentation.} Table~\ref{tab:part-results} shows that SegPFN achieves the highest average across one- and five-shot settings (57.15), exceeding FoRIS (53.19) by 3.96 points. It also achieves the highest two-dataset average in both the one-shot (51.97 vs. 49.17 for FoRIS, +2.80) and five-shot (62.32 vs. 59.22 for FROST, +3.10) settings. At the dataset level, SegPFN leads on both datasets in one-shot and on PACO-Part in five-shot (57.55); FROST narrowly leads on PASCAL-Part (67.53 vs. 67.09).

\paragraph{Personalized segmentation.} Table~\ref{tab:personalized} shows that SegPFN achieves the highest one-shot average across the three datasets (59.13), exceeding FROST (53.93) by 5.20 points. It leads on PerMIS (71.76; +4.59 over INSID3) and DAVIS-2017-ICS (55.92; +5.74 over FROST), while Matcher achieves the highest score on CUHK-SYSU-ICS (50.02), narrowly ahead of SegPFN (49.71).

\paragraph{Discussion.} SegPFN achieves the highest average mIoU in 4 of 5 task--shot settings and the highest Overall mIoU (60.32 vs. 55.62 for FROST). Its largest gains occur in part and personalized segmentation, which require distinguishing target regions from semantically related parts or instances. SegPFN's ability to jointly leverage positive and negative reference examples may help capture these distinctions. Overall, these results demonstrate that a pretrained tabular PFN can support ICS across diverse segmentation tasks and image domains without segmentation-specific training, displaying particular strengths in fine-grained target selection. 

\subsection{Understanding SegPFN's Advantages}
\label{sec:advantages}

\begin{figure*}[!t]
\centering
\includegraphics[width=0.96\textwidth]{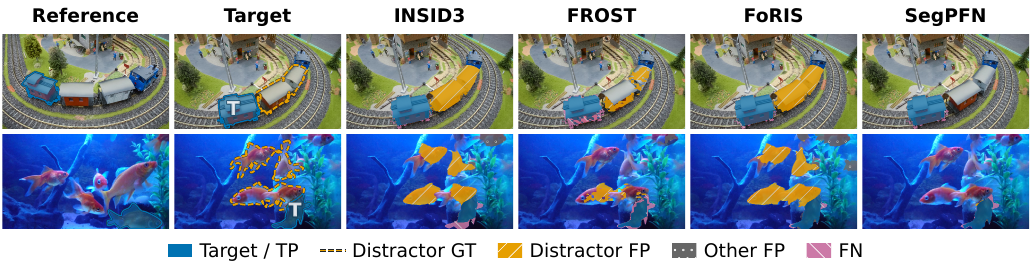}
\captionsetup{skip=3pt}
\caption{\textbf{Personalized segmentation with same-category distractors.} Reference masks specify the target instance; all methods use identical reference--target image pairs and reference masks. Mask legend: \segmasktp~target mask / true positive (TP); \segmaskgt~distractor ground truth (GT); \segmaskhardfp~false positive (FP) on a distractor; \segmaskotherfp~other FP; \segmaskfn~false negative (FN).}
\label{fig:personalized-distractors}
\end{figure*}

\paragraph{Better discrimination between instances of the same category.}
SegPFN achieves the highest average mIoU across the three
personalized segmentation benchmarks
(Table~\ref{tab:personalized}). This motivates the hypothesis
that SegPFN better distinguishes the requested instance from
\emph{distractors}, i.e., non-target instances of the same
semantic category. Figure~\ref{fig:personalized-distractors}
illustrates this challenge: a model may recognize the requested
category but also predict other instances as foreground.
Let FP and TP denote false-positive and true-positive pixel counts. We compare SegPFN with INSID3,
FROST, and FoRIS using the \emph{FP ratio},
$100\sum\mathrm{FP}/\sum(\mathrm{TP}+\mathrm{FP})$, pooled
across tasks within each dataset, and the \emph{distractor
positive rate} (DPR) on DAVIS-2017-ICS. DPR measures the
percentage of annotated distractor pixels in the target
image predicted as foreground, averaged over tasks with
nonempty distractor annotations.

\begin{table}[!htbp]
\centering
\captionsetup{skip=6pt}
\caption{False-positive analysis of one-shot personalized segmentation (\%, $\downarrow$). FP ratio uses all tasks. DPR uses the 53 DAVIS-2017-ICS tasks with annotated distractors. Bold/underline mark the lowest/second-lowest values.}
\label{tab:false-positives}
\small
\setlength{\tabcolsep}{1.4pt}
\renewcommand{\arraystretch}{1.20}
\begin{tabular*}{\columnwidth}{@{\extracolsep{\fill}}lcccc@{}}
\toprule
\multirow{2}{*}{Method} & \multicolumn{1}{c}{PerMIS} & \multicolumn{1}{c}{CUHK-SYSU-ICS} & \multicolumn{2}{c}{DAVIS-2017-ICS} \\
\cmidrule(lr){2-2}\cmidrule(lr){3-3}\cmidrule(l){4-5}
& \multicolumn{1}{c}{FP ratio $\downarrow$} & \multicolumn{1}{c}{FP ratio $\downarrow$} & \multicolumn{1}{c}{FP ratio $\downarrow$} & \multicolumn{1}{c@{}}{DPR $\downarrow$} \\
\midrule
INSID3~\cite{cuttano2026insid3} & \textbf{27.51} & \underline{45.03} & 53.52 & 37.72 \\
FROST~\cite{park2026frost} & 30.91 & 50.86 & \underline{40.18} & \underline{17.13} \\
FoRIS~\cite{hu2026foris} & 37.33 & 50.71 & 58.84 & 55.20 \\
\midrule
SegPFN & \underline{27.69} & \textbf{39.29} & \textbf{31.12} & \textbf{13.56} \\
\bottomrule
\end{tabular*}
\end{table}


Among the evaluated methods, SegPFN achieves the lowest
FP ratios on DAVIS-2017-ICS and CUHK-SYSU-ICS
(Table~\ref{tab:false-positives}). Its scores of 31.12\% and 39.29\% are lower than those of the strongest baseline in each column by 9.06 and 5.74 percentage points, respectively.
On PerMIS, INSID3 has a marginally lower FP ratio
(27.51\% versus 27.69\%). SegPFN also achieves the lowest
DPR on DAVIS-2017-ICS (13.56\% versus 17.13\% for FROST),
supporting its ability to exclude other instances of the
same category on this benchmark.

\paragraph{Hard-negative reference patches improve target discrimination through tabular in-context learning.}
We hypothesize that hard-negative reference patches play
a key role in this advantage by providing informative
negative examples during tabular in-context learning.
We define \emph{hard-negative reference patches} as
background-labeled patches from semantically related
non-target regions in the reference image.
In DAVIS-2017-ICS, these regions correspond to other
instances of the same semantic category as the requested
instance. In PASCAL-Part and PACO-Part, they correspond
to other annotated parts of the same parent object as
the requested part. Target and ignored pixels are
excluded from these regions.

To test this hypothesis, we replace these patches in
the sampled PFN context with other background patches,
keeping the foreground/background context budgets and
prediction refinement unchanged
(Sec.~\ref{sec:refinement}).
We evaluate all 53 DAVIS-2017-ICS tasks and randomly
sampled subsets of 500 tasks for each part dataset and
shot setting. Alongside mIoU, we report the
\emph{non-target positive rate} (NTPR): the percentage
of pixels in annotated semantically related non-target regions
of the target image predicted as foreground, averaged over the
$n_{\mathrm{NT}}$ tasks with nonempty non-target \mbox{annotations}.
NTPR is equivalent to DPR on DAVIS-2017-ICS.

Replacing hard-negative reference patches increases NTPR
and reduces mIoU in every evaluated setting
(Table~\ref{tab:reference-negatives}).
On DAVIS-2017-ICS, NTPR rises from 13.56\% to 66.44\%,
while mIoU falls from 55.92\% to 33.24\%, a decline of
22.68 points. Across the part settings, NTPR increases by
17.14--22.76 points and mIoU decreases by 6.48--13.55 points,
with larger mIoU drops in five-shot on both datasets.
Figure~\ref{fig:reference-negatives} illustrates the
resulting false positives on distractors and non-target
parts. These results support the hypothesis that hard-negative reference examples help SegPFN distinguish the target from semantically related non-target regions and improve segmentation accuracy.

\begingroup\captionsetup{hypcap=false}\begin{table}[!t]
\centering
\begin{minipage}{\linewidth}
\centering
\includegraphics[width=\linewidth]{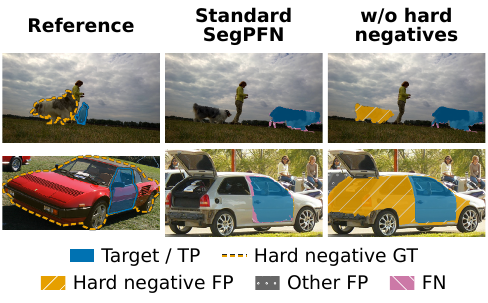}
\captionsetup{skip=5pt}
\captionof{figure}{\textbf{Hard-negative reference patches guide target selection.} Replacing hard-negative reference patches with other background patches increases false positives in both the dog and car-part examples. Mask legend: \segmasktp~target mask / true positive (TP); \segmaskgt~hard-negative ground truth (GT); \segmaskhardfp~false positive (FP) on a hard-negative region; \segmaskotherfp~other FP; \segmaskfn~false negative (FN).}
\label{fig:reference-negatives}
\end{minipage}
\par\vspace{8pt}
\captionsetup{skip=6pt}
\caption{Effect of hard-negative reference-patch replacement. Std.: standard sampling; Repl.: hard-negative replacement. NTPR ($\downarrow$) and mIoU ($\uparrow$) are percentages; bold marks the better paired value. NTPR uses the $n_{\mathrm{NT}}$ eligible tasks, while mIoU uses all $N$ tasks and the same aggregation procedure as the main results. Part results use randomly sampled 500-task benchmark subsets.}
\label{tab:reference-negatives}
\footnotesize
\setlength{\tabcolsep}{1.5pt}
\renewcommand{\arraystretch}{1.60}
\begin{tabular*}{\columnwidth}{@{\extracolsep{\fill}}lcrrrrr@{}}
\toprule
\multirow{2}{*}{Dataset} & \multirow{2}{*}{Shot} & \multirow{2}{*}{$n_{\mathrm{NT}}/N$} & \multicolumn{2}{c}{NTPR $\downarrow$} & \multicolumn{2}{c@{}}{mIoU $\uparrow$} \\
\cmidrule(lr){4-5}\cmidrule(l){6-7}
& & & Std. & Repl. & Std. & Repl. \\
\midrule
DAVIS-2017-ICS & 1 & 53/53 & \textbf{13.56} & 66.44 & \textbf{55.92} & 33.24 \\
\addlinespace[5pt]
\multirow{2}{*}{PASCAL-Part} & 1 & 410/500 & \textbf{11.03} & 28.17 & \textbf{55.89} & 48.17 \\
& 5 & 410/500 & \textbf{8.06} & 30.82 & \textbf{67.16} & 53.61 \\
\addlinespace[5pt]
\multirow{2}{*}{PACO-Part} & 1 & 449/500 & \textbf{15.68} & 33.59 & \textbf{45.52} & 39.04 \\
& 5 & 445/500 & \textbf{12.23} & 31.89 & \textbf{56.54} & 46.17 \\
\bottomrule
\end{tabular*}
\end{table}\endgroup


\paragraph{More effective use of multiple references through joint conditioning.}
SegPFN trails FoRIS in the one-shot semantic average but
leads in five-shot (Table~\ref{tab:semantic-results}).
We hypothesize that jointly using multiple references
through tabular in-context learning contributes to this
advantage. We test this hypothesis by comparing joint five-shot
SegPFN with a $5\times$1-shot ensemble.
All methods use DINOv3-L and the same randomly sampled
benchmark subsets of up to $1{,}000$ evaluation tasks per dataset.

The two SegPFN variants share the five references,
reference-fitted PCA projection
(Sec.~\ref{sec:features}), and refinement
(Sec.~\ref{sec:refinement}).
The ensemble produces one probability map for each reference, using
up to $1{,}024$ patches per PFN context, and averages the
five maps before refinement.
Joint conditioning (Sec.~\ref{sec:features})
instead draws each PFN context from all five references, with up to $1{,}024$ patches in total.

\begin{table}[!htbp]
\centering
\caption{Five-shot reference-conditioning comparison on randomly sampled benchmark subsets (mIoU in \%, $\uparrow$). Semantic and Part report averages across six and two datasets, respectively; Avg.~(Sem./Part) equally weights the two task means. All methods use DINOv3-L and identical evaluation tasks. SegPFN variants share the feature projection and refinement; context budgets are detailed in the text. Best scores are bold.}
\label{tab:joint-conditioning}
\footnotesize
\setlength{\tabcolsep}{3pt}
\renewcommand{\arraystretch}{1.08}
\begin{tabular*}{\columnwidth}{@{\extracolsep{\fill}}lrrr@{}}
\toprule
Method & \multicolumn{1}{c}{Semantic} & \multicolumn{1}{c}{Part} & \multicolumn{1}{c}{\shortstack{Avg.\\(Sem./Part)}} \\
\midrule
FoRIS~\cite{hu2026foris} & 67.89 & 57.58 & 62.73 \\
INSID3~\cite{cuttano2026insid3} & 63.41 & 53.64 & 58.52 \\
FROST~\cite{park2026frost} & 64.74 & 59.49 & 62.12 \\
\midrule
\shortstack[l]{SegPFN\\{\scriptsize ($5\times$1-shot ensemble)}} & 66.27 & 57.07 & 61.67 \\
\addlinespace[2pt]
\shortstack[l]{\textbf{SegPFN}\\{\scriptsize (joint 5-shot)}} & \textbf{68.96} & \textbf{62.87} & \textbf{65.91} \\
\bottomrule
\end{tabular*}
\end{table}

The $5\times$1-shot ensemble achieves strong performance, with an average mIoU of 61.67, outperforming INSID3 (58.52)
and approaching FoRIS (62.73) and FROST (62.12).
Nevertheless, joint five-shot SegPFN raises this average
to 65.91, a further gain of 4.24 points
(Table~\ref{tab:joint-conditioning}).
Joint conditioning improves semantic and part segmentation mIoU by 2.69 and 5.80 points, respectively, despite its
smaller maximum total reference-patch budget.
These results support our hypothesis that joint conditioning during PFN inference contributes to SegPFN’s five-shot advantage.

\newpage
\subsection{Ablation Studies}
\label{sec:ablations}

\paragraph{Contribution of PFN inference.}
Does pretrained PFN inference improve segmentation beyond conventional classification methods? To investigate this, we replace TabPFN with $k$-nearest neighbors (kNN), linear ridge regression,
ridge regression with a radial basis function (RBF) kernel, or logistic regression, keeping the frozen
visual features, sampled reference contexts, and refinement
fixed (Sec.~\ref{sec:refinement}).
We evaluate randomly sampled benchmark subsets of up to $1{,}000$
evaluation tasks per dataset and shot setting. For each conventional classifier, we select the tested configuration with the highest Overall score on these same subsets, without using an independent held-out set for selection. Search grids and subset details are provided
in the supplementary material. TabPFN leads all five task--shot settings
(Table~\ref{tab:inference-models}), achieving 60.67 Overall
versus 58.24 for logistic regression, the strongest
conventional classifier. This 2.43-point gain supports
a contribution from PFN inference beyond the shared
dense classification formulation.

\begin{table}[!htbp]
\centering
\caption{Inference-model comparison on randomly sampled benchmark subsets (mIoU in \%, $\uparrow$). Conventional predictors use their best tested configuration on the reported subsets. Task means equally weight six semantic, two part, or three personalized datasets. Overall equally weights the three tasks, averaging shots within Semantic and Part. All predictors share refinement. Best scores are bold.}
\label{tab:inference-models}
\footnotesize
\setlength{\tabcolsep}{2pt}
\renewcommand{\arraystretch}{1.08}
\begin{tabular*}{\columnwidth}{@{\extracolsep{\fill}}lrrrrrr@{}}
\toprule
\multirow{2}{*}{Predictor} & \multicolumn{2}{c}{Semantic} & \multicolumn{2}{c}{Part} & \multicolumn{1}{c}{Personalized} & \multirow{2}{*}{Overall} \\
\cmidrule(lr){2-3}\cmidrule(lr){4-5}\cmidrule(lr){6-6}
& \multicolumn{1}{c}{1-shot} & \multicolumn{1}{c}{5-shot} & \multicolumn{1}{c}{1-shot} & \multicolumn{1}{c}{5-shot} & \multicolumn{1}{c}{1-shot} & \\
\midrule
kNN & 50.87 & 60.28 & 43.22 & 54.90 & 56.53 & 53.72 \\
RBF ridge & 57.88 & 66.05 & 45.54 & 53.99 & 48.52 & 53.42 \\
Linear ridge & 57.62 & 65.66 & 45.09 & 53.78 & 48.44 & 53.17 \\
Logistic & 56.60 & 67.67 & 46.97 & 61.51 & 58.34 & 58.24 \\
\midrule
\shortstack[l]{\textbf{TabPFN 3.5}\\{\scriptsize (SegPFN)}} & \textbf{61.11} & \textbf{68.96} & \textbf{52.81} & \textbf{62.87} & \textbf{59.13} & \textbf{60.67} \\
\bottomrule
\end{tabular*}
\end{table}

\newpage
\paragraph{Contribution of prediction refinement.}
Does prediction refinement further improve segmentation accuracy? To assess this, we progressively add cluster smoothing and
reference-guided correction to the PFN baseline
(Sec.~\ref{sec:refinement}).
All variants retain min--max normalization and CRF and
follow the main-benchmark evaluation protocol
(Sec.~\ref{sec:experiments}).

\begin{table}[!htbp]
\centering
\caption{Cumulative refinement ablation under the main-benchmark protocol (mIoU in \%, $\uparrow$). All rows retain min--max normalization and CRF. Semantic and Part average the one-shot and five-shot results across six and two datasets, respectively; Personalized averages the one-shot results across three datasets. Overall equally weights the three task means. Changes below scores are relative to the preceding row, in percentage points, computed from displayed scores. Best scores are bold.}
\label{tab:ablations}
\footnotesize
\setlength{\tabcolsep}{2pt}
\renewcommand{\arraystretch}{1.08}
\newcommand{\segrefcell}[2]{\shortstack{#1\\[-1pt]{\scriptsize $(#2)$}}}
\begin{tabular*}{\columnwidth}{@{\extracolsep{\fill}}lcccc@{}}
\toprule
Configuration & Semantic & Part & Personalized & Overall \\
\midrule
PFN baseline & 61.69 & 53.18 & 59.01 & 57.96 \\
\addlinespace[3pt]
\shortstack[l]{+ Cluster\\\hphantom{+ }smoothing} & \segrefcell{63.19}{+1.50} & \segrefcell{56.13}{+2.95} & \segrefcell{58.18}{-0.83} & \segrefcell{59.17}{+1.21} \\
\addlinespace[3pt]
\shortstack[l]{+ Reference-guided\\\hphantom{+ }correction} & \segrefcell{\textbf{64.67}}{+1.48} & \segrefcell{\textbf{57.15}}{+1.02} & \segrefcell{\textbf{59.13}}{+0.95} & \segrefcell{\textbf{60.32}}{+1.15} \\
\bottomrule
\end{tabular*}
\end{table}

Cluster smoothing raises Overall mIoU from 57.96 to 59.17, improving semantic and part segmentation while reducing personalized segmentation mIoU from 59.01 to 58.18. Reference-guided correction further raises Overall to 60.32, a cumulative 2.36-point gain (Table~\ref{tab:ablations}). Detailed results by dataset and shot setting are provided in the supplementary material.


\section{Conclusion and Limitations}
\label{sec:conclusion}

We introduced SegPFN, a training-free framework that reformulates in-context segmentation as tabular classification using pretrained visual and tabular foundation models. Experiments across semantic, part, and personalized segmentation demonstrate that a frozen PFN pretrained on tabular tasks provides a competitive alternative to conventional feature-matching and aggregation approaches. Our analyses further show that hard-negative reference patches reduce false positives on semantically related non-target regions, while joint reference conditioning improves five-shot performance over an ensemble of one-shot predictions. These findings support transferring tabular in-context learning to visual segmentation without segmentation-specific training.

Three limitations remain. First, SegPFN still relies on manually designed components for feature processing and prediction refinement. Future work could explore more integrated approaches that reduce reliance on these components. Second, its computational efficiency and scalability to larger images and reference sets remain to be evaluated. Third, its performance depends on the quality of frozen visual representations, which may not capture all discriminative information needed for fine-grained segmentation.
\clearpage
{
    \small
    \bibliographystyle{ieeenat_fullname}
    \bibliography{main}
}

\clearpage
\appendix
\setcounter{page}{1}\renewcommand{\thepage}{S\arabic{page}}
\maketitlesupplementary
\raggedbottom
\captionsetup[table]{position=top,skip=5pt}
\setcounter{dbltopnumber}{1}

\section*{Overview of the Supplementary Material}
In this supplement, we provide implementation details for SegPFN, report additional experiments and analyses, and describe how we constructed our two new benchmarks.

\noindent Specifically:

\begin{itemize}[leftmargin=*,topsep=.3em,itemsep=.35em,parsep=0pt]
\item \textbf{Implementation and evaluation.} In Sec.~\ref{sec:supp-experiment-protocols}, we describe how we sample reference patches, refine predicted masks, and aggregate benchmark scores.

\item \textbf{Personalized benchmarks.} In Sec.~\ref{sec:supp-personalized-construction}, we describe task construction and identity checks for DAVIS-2017-ICS and CUHK-SYSU-ICS, then discuss uncertainty when tasks share images or videos.

\item \textbf{Negative reference patches.} In Sec.~\ref{sec:supp-negatives}, we vary the fraction of hard negatives retained by four methods and measure segmentation accuracy and false positives on related non-target regions.

\item \textbf{Reference efficiency.} In Sec.~\ref{sec:supp-references}, we compare four methods across reference counts and evaluate SegPFN against five-shot baselines when it uses fewer annotated references.

\item \textbf{Conventional classifiers.} In Sec.~\ref{sec:supp-classifiers}, we report the full hyperparameter comparison and dataset-level results for the four conventional classifiers evaluated against SegPFN.

\item \textbf{Prediction refinement.} In Sec.~\ref{sec:supp-component-ablation}, we report how cluster smoothing and reference-guided correction affect individual datasets in the main DINOv3-L evaluation.

\item \textbf{Robustness across backbone sizes.} In Sec.~\ref{sec:supp-backbone}, we test whether SegPFN's advantages persist with smaller DINOv3 backbones and examine how backbone size affects refinement.

\item \textbf{Inference efficiency.} In Sec.~\ref{sec:supp-efficiency}, we report one- and five-shot inference times for four training-free methods using cached visual features.

\item \textbf{Conceptual contribution.} In Sec.~\ref{sec:supp-scope}, we explain how SegPFN reformulates segmentation as tabular in-context classification and what the experiments establish about this transfer.
\end{itemize}

\section{Implementation and Evaluation Details}
\label{sec:supp-experiment-protocols}
\subsection{SegPFN inference settings}
\label{sec:supp-segpfn-inference-settings}

\paragraph{Patch features and reference views.}
By default, SegPFN uses both raw and positionally debiased DINOv3-L patch features, with positional debiasing performed as in INSID3~\cite{cuttano2026insid3}.
We fit PCA separately to the two representations using only patch features from the original reference images. Each projection maps the corresponding reference and target features to 128 dimensions (Sec.~\ref{sec:features}).
For one-shot inference, we use the original annotated reference together with four augmented views. Each view is generated by applying color jitter (Table~\ref{tab:supp-segpfn-inference-settings}) followed by a distinct spatial transform: a horizontal flip, a $+15^\circ$ rotation, a $-15^\circ$ rotation, or a translation to the right by 10\% of the image width and down by 10\% of its height. We apply the same spatial transform to the image and its mask.
For five-shot inference, we use the five original reference images. For any multi-shot setting ($K>1$), we append each patch's mean RGB values $(\bar r,\bar g,\bar b)$ over valid pixels and its normalized coordinates $(x,y)$ to the PCA features, yielding a 133-dimensional PFN input.

\paragraph{PFN ensemble.}
We use an ensemble of four frozen TabPFN-3.5 estimators: two use raw features and two use positionally debiased features.
Each estimator independently samples labeled reference patches to form its context, then predicts a foreground probability for each target patch.
To produce the final mask, we average the four probability maps and then apply cluster smoothing, a reference-guided correction adapted from FoRIS~\cite{hu2026foris}, and CRF boundary refinement (Sec.~\ref{sec:refinement}).

\paragraph{Hyperparameter selection.}
We select SegPFN's inference hyperparameters using three-fold cross-validation on the COCO-$20^i$ training split, following the validation protocol of INSID3~\cite{cuttano2026insid3}.
Table~\ref{tab:supp-segpfn-inference-settings} lists the final hyperparameters for one-shot and five-shot inference. Within each shot setting, we use the same values across all datasets and segmentation tasks.

\begin{table}[!htbp]
    \centering
    \caption{Feature and PFN settings for SegPFN inference.}
    \label{tab:supp-segpfn-inference-settings}
    \footnotesize
    \begin{tabular}{@{}p{.315\columnwidth}@{\hspace{.03\columnwidth}}p{.655\columnwidth}@{}}
        \toprule
        Setting & Value \\
        \midrule
        Positional debiasing & INSID3 settings~\cite{cuttano2026insid3} \\
        Reference views & \raggedright 1 original + 4 augmented (one shot)\newline 5 originals (five shots) \tabularnewline
        One-shot color jitter & \raggedright Brightness, contrast, and saturation factors $[0.8,1.2]$\newline Hue offset $[-0.05,0.05]$ \tabularnewline
        PCA dimension & 128 per representation \\
        PFN input dimension & \raggedright 128 PCA features (one shot)\newline $128+3+2=133$ (two or more shots)\newline PCA + mean RGB + normalized $(x,y)$ \tabularnewline
        PFN ensemble & 2 raw + 2 debiased estimators \\
        Reference context & \raggedright Up to 1,024 labeled patches per estimator across all references \tabularnewline
        \bottomrule
    \end{tabular}
\end{table}

\subsection{Evaluation protocols and metric aggregation}
\label{sec:supp-cohorts}

\paragraph{Main benchmark protocols.}
We evaluate SegPFN on the full set of tasks for each benchmark (Table~\ref{tab:evaluation-protocols}). We report one-shot and five-shot results for semantic and part segmentation, and one-shot results for personalized segmentation. We evaluate SegIC on all three personalized benchmarks using the PerMIS inference settings from the INSID3 study~\cite{cuttano2026insid3}.

\begin{table}[!htbp]
\centering
\caption{Number of tasks used to evaluate SegPFN in the main comparisons. DAVIS-2017-ICS and CUHK-SYSU-ICS each use a fixed list of one-shot tasks.}
\label{tab:evaluation-protocols}
\small
\setlength{\tabcolsep}{3pt}
\begin{tabular*}{\columnwidth}{@{\extracolsep{\fill}}lrr@{}}
\toprule
Dataset & One-shot tasks & Five-shot tasks \\
\midrule
\multicolumn{3}{@{}l}{\textit{Semantic segmentation}} \\
COCO-$20^i$ & 4,000 & 4,000 \\
LVIS-$92^i$ & 23,000 & 23,000 \\
ISIC & 600 & 600 \\
Lung & 600 & 600 \\
iSAID & 10,063 & 10,063 \\
SUIM & 3,859 & 3,859 \\
\addlinespace[2pt]
\multicolumn{3}{@{}l}{\textit{Part segmentation}} \\
PASCAL-Part & 8,459 & 8,459 \\
PACO-Part & 10,000 & 10,000 \\
\addlinespace[2pt]
\multicolumn{3}{@{}l}{\textit{Personalized segmentation}} \\
PerMIS & 432 & \textemdash \\
DAVIS-2017-ICS & 53 & \textemdash \\
CUHK-SYSU-ICS & 151 & \textemdash \\
\bottomrule
\end{tabular*}
\end{table}

\paragraph{Subsets for controlled comparisons.}
Table~\ref{tab:supp-backbone-cohort} lists the number of one-shot and five-shot tasks per dataset for the classifier and backbone comparisons. For the backbone comparison (Sec.~\ref{sec:supp-backbone}), we sample semantic and part tasks from the DINOv3-L evaluation pool (seed~17) and retain all personalized tasks. We use all $7{,}200$ five-shot tasks in these subsets to compare joint conditioning with the $5\times$1-shot ensemble (Table~\ref{tab:joint-conditioning}). Both strategies use the same target image and the same five annotated references in each task. For the hard-negative analysis (Sec.~\ref{sec:supp-negatives}), we select tasks separately, using all 53 one-shot tasks from DAVIS-2017-ICS and 500 tasks per shot setting from each of PASCAL-Part and PACO-Part.

\begin{table}[!htbp]
\centering
\caption{Evaluation subsets for the classifier and backbone comparisons. The five-reference comparison uses the five-shot column. Personalized segmentation uses one shot only.}
\label{tab:supp-backbone-cohort}
\footnotesize
\setlength{\tabcolsep}{4pt}
\renewcommand{\arraystretch}{1.08}
\begin{tabular*}{\columnwidth}{@{\extracolsep{\fill}}lrr@{}}
\toprule
Dataset & 1-shot & 5-shot \\
\midrule
\multicolumn{3}{@{}l}{\textit{Semantic segmentation}} \\
COCO-$20^i$ & 1,000 & 1,000 \\
LVIS-$92^i$ & 1,000 & 1,000 \\
ISIC & 600 & 600 \\
Lung & 600 & 600 \\
iSAID & 1,000 & 1,000 \\
SUIM & 1,000 & 1,000 \\
\addlinespace[2pt]
\multicolumn{3}{@{}l}{\textit{Part segmentation}} \\
PASCAL-Part & 1,000 & 1,000 \\
PACO-Part & 1,000 & 1,000 \\
\addlinespace[2pt]
\multicolumn{3}{@{}l}{\textit{Personalized segmentation}} \\
PerMIS & 432 & \textemdash \\
DAVIS-2017-ICS & 53 & \textemdash \\
CUHK-SYSU-ICS & 151 & \textemdash \\
\midrule
Total & 7,836 & 7,200 \\
\bottomrule
\end{tabular*}
\end{table}

\paragraph{Metric aggregation.} For semantic and part segmentation, we compute dataset mIoU by averaging the class-wise IoUs. For personalized segmentation, we compute foreground IoU by pooling the intersection and union counts across all tasks within each dataset.

The \emph{Total Avg.} averages the one- and five-shot scores for Semantic and Part, and uses the one-shot score for Personalized. The \emph{Overall} score is the unweighted mean of the three task-type scores:
\[
\mathrm{Overall} =
\frac{\mathrm{Semantic}+\mathrm{Part}+\mathrm{Personalized}}{3}.
\]

\section{Benchmark Construction and Reliability}
\label{sec:supp-personalized-construction}
This section describes how we constructed our two personalized segmentation benchmarks, DAVIS-2017-ICS and CUHK-SYSU-ICS, and examines the stability of method comparisons on each. Table~\ref{tab:supp-ics-size} shows how many reference--target image pairs each benchmark contains and how many one-shot tasks those pairs yield.

\begin{table}[!htbp]
\centering
\caption{\textbf{Personalized benchmark sizes.} Each one-shot task selects one instance in an ordered reference--target pair. We count each pair once across tasks.}
\label{tab:supp-ics-size}
\small
\renewcommand{\arraystretch}{1.12}
\begin{tabular*}{\columnwidth}{@{\extracolsep{\fill}}lcc@{}}
\toprule
\textbf{Benchmark} & \textbf{Tasks} & \textbf{Reference--target pairs} \\
\midrule
DAVIS-2017-ICS & 53 & 20 \\
CUHK-SYSU-ICS & 151 & 113 \\
\bottomrule
\end{tabular*}
\end{table}

\subsection{DAVIS-2017-ICS}
\begin{figure}[!htbp]
\centering
\includegraphics[width=\columnwidth]{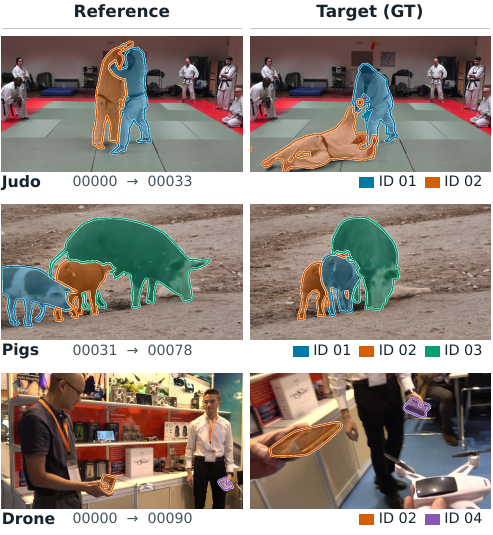}
\caption{\textbf{Selecting instances on fixed DAVIS frame pairs.} The same color identifies an athlete, pig, or phone across the reference and target frames. Each colored instance defines a separate one-shot task with its own binary reference mask.}
\label{fig:supp-davis-target-switch}
\end{figure}

We screened the 90 videos in the DAVIS 2017 training and validation splits~\cite{ponttuset2017davis} and identified 29 candidates with separate masks for multiple instances of the same category. We sampled up to 12 frames per candidate video (one from each of ten intervals plus the first and last frames). For a given instance, we excluded a frame pair if \textcircled{\scriptsize 1}~the reference-to-target gap was under 24 frames or under 25\% of the sequence's frame span, \textcircled{\scriptsize 2}~we could not identify the same instance in both frames on manual review of the masks and IDs, or \textcircled{\scriptsize 3}~the target frame contained no other annotated instance of the same category. We favored pairs showing a change in the instance's position or pose. After manual review, we retained one frame pair from each of 20 videos (40 images in total).
These pairs yielded 53 one-shot tasks across nine categories, with 35 tasks from training videos and 18 from validation videos. Figure~\ref{fig:supp-davis-target-switch} shows example frame pairs in which colors distinguish instances of the same category. For a fixed pair, each task uses a different reference mask to specify which instance to segment in the target frame.

\subsection{CUHK-SYSU-ICS}
\begin{figure}[!htbp]
\centering
\includegraphics[width=\columnwidth]{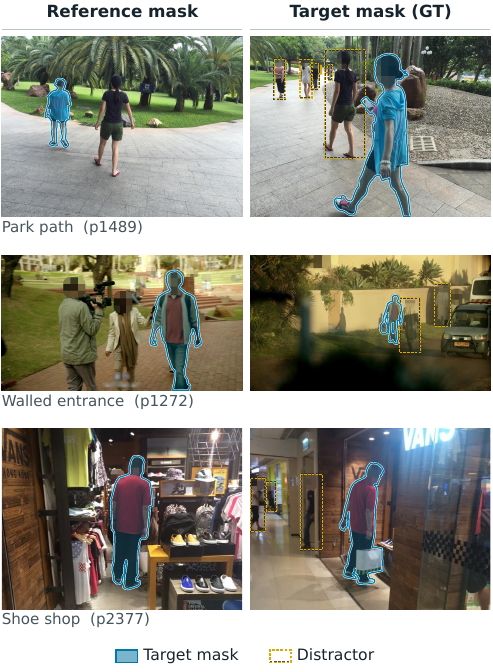}
\caption{\textbf{Following one person across CUHK-SYSU images.} Blue masks mark the requested person in the reference (left) and target ground truth (right). Dashed yellow boxes mark other people. For display only, we pixelate visible faces for privacy.}
\label{fig:supp-cuhk-person-identity}
\end{figure}

We began with the CUHK-SYSU training split, which contains 11,206 images and 5,532 labeled identities~\cite{xiao2017personsearch}. We matched its person bounding boxes to manually annotated polygons from maskPS~\cite{zheng2020segmentation}. For a given person, an ordered pair of distinct images qualified if \textcircled{\scriptsize 1}~both images contained a valid mask for that person and \textcircled{\scriptsize 2}~the target image contained a bounding box for at least one other person. These criteria yielded 397 candidate reference--target assignments across 195 identities. During manual review, we excluded a pair if the two views of the person were nearly identical or if the person's outline was too ambiguous for reliable segmentation. We retained 113 distinct ordered image pairs and verified the reference and target masks against the original polygons.
These pairs yielded 151 one-shot tasks, with some image pairs shared across identities. Figure~\ref{fig:supp-cuhk-person-identity} shows three examples.

\subsection{Statistical stability of method comparisons}
\label{sec:supp-personalized-stability}
\begin{table}[!b]
\centering
\caption{\textbf{Stability of one-shot pooled foreground IoU.} Method scores are percentages. $\Delta$ is the paired difference (SegPFN minus baseline) in percentage points (pp). We report 95\% bootstrap percentile confidence intervals (CIs). The last row gives the range of $\Delta$ after omitting each group in turn.}
\label{tab:supp-personalized-stability}
\small
\setlength{\tabcolsep}{1.5pt}
\renewcommand{\arraystretch}{1.12}
\begin{tabular*}{\columnwidth}{@{\extracolsep{\fill}}l rc rc@{}}
\toprule
& \multicolumn{2}{c}{\textbf{DAVIS-2017-ICS}} & \multicolumn{2}{c}{\textbf{CUHK-SYSU-ICS}} \\
Best baseline & \multicolumn{2}{c}{FROST} & \multicolumn{2}{c}{Matcher} \\
\cmidrule(lr){2-3}\cmidrule(l){4-5}
& Est. & 95\% CI & Est. & 95\% CI \\
\midrule
SegPFN & $55.92$ & $[46.31, 67.33]$ & $49.71$ & $[43.17, 56.20]$ \\
Baseline & $50.18$ & $[41.43, 59.63]$ & $50.02$ & $[45.11, 54.90]$ \\
\addlinespace[2pt]
$\Delta$ (pp) & $+5.74$ & $[0.60, 11.12]$ & $-0.31$ & $[-6.16, 5.55]$ \\
Omit one group & \multicolumn{2}{c}{$[+4.29, +7.10]$} & \multicolumn{2}{c}{$[-1.57, +0.66]$} \\
\bottomrule
\end{tabular*}
\end{table}

\begin{table*}[tp]
\centering
\caption{\textbf{Hard-negative retention across datasets and shot settings.}
The left columns report foreground mIoU and the right columns report the non-target positive rate (NTPR).
Retention percentages refer to each method's own pool of hard-negative reference patches.
Counts in parentheses give the number of tasks with annotated non-target pixels used for NTPR out of the total number evaluated for mIoU.}
\label{tab:supp-hard-negative-retention}
\small
\setlength{\tabcolsep}{2pt}
\renewcommand{\arraystretch}{1.04}
\begin{tabular*}{\linewidth}{@{\extracolsep{\fill}}llrrrrr@{\hspace{12pt}}rrrrr@{}}
\toprule
Dataset / shot & Method & \multicolumn{5}{c}{Foreground mIoU (\%) $\uparrow$} & \multicolumn{5}{c}{Non-target positive rate (\%) $\downarrow$} \\
\cmidrule(lr){3-7}\cmidrule(l){8-12}
\multicolumn{2}{@{}l}{\textit{Hard negatives retained (\%)}} & 0 & 10 & 25 & 50 & 100 & 0 & 10 & 25 & 50 & 100 \\
\midrule
\multirow{4}{*}{\shortstack[l]{\textbf{DAVIS-2017-ICS}\\1-shot (53/53)}} & \textbf{SegPFN} & 33.24 & 56.42 & 57.45 & 57.18 & 55.92 & 66.44 & 24.27 & 16.67 & 14.75 & 13.56 \\
 & FoRIS & 31.12 & 32.30 & 35.07 & 38.26 & 38.68 & 73.48 & 71.04 & 66.27 & 60.17 & 55.20 \\
 & FROST & 50.34 & 51.10 & 50.88 & 50.68 & 50.18 & 10.47 & 12.32 & 13.88 & 15.25 & 17.13 \\
 & INSID3 & 30.18 & 39.60 & 40.02 & 39.96 & 38.75 & 69.71 & 45.27 & 43.07 & 40.72 & 37.72 \\
\midrule
\multirow{4}{*}{\shortstack[l]{\textbf{PASCAL-Part}\\1-shot (410/500)}} & \textbf{SegPFN} & 48.17 & 55.65 & 56.33 & 56.04 & 55.89 & 28.17 & 13.15 & 11.85 & 11.32 & 11.03 \\
 & FoRIS & 50.66 & 52.64 & 53.70 & 53.96 & 54.04 & 17.33 & 13.44 & 11.32 & 10.35 & 9.99 \\
 & FROST & 52.66 & 52.11 & 51.57 & 51.25 & 51.18 & 13.53 & 14.87 & 15.35 & 15.57 & 15.17 \\
 & INSID3 & 43.86 & 48.61 & 48.47 & 48.53 & 48.98 & 23.40 & 11.50 & 10.92 & 10.72 & 10.27 \\
\midrule
\multirow{4}{*}{\shortstack[l]{\textbf{PASCAL-Part}\\5-shot (410/500)}} & \textbf{SegPFN} & 53.61 & 65.55 & 66.31 & 66.81 & 67.16 & 30.82 & 11.71 & 9.86 & 8.93 & 8.06 \\
 & FoRIS & 58.96 & 61.01 & 62.07 & 62.48 & 62.60 & 16.11 & 12.40 & 10.44 & 9.18 & 8.36 \\
 & FROST & 65.39 & 65.62 & 65.73 & 65.86 & 65.93 & 10.97 & 10.63 & 10.27 & 9.87 & 9.04 \\
 & INSID3 & 49.43 & 57.59 & 57.39 & 57.42 & 57.42 & 30.02 & 13.75 & 13.05 & 12.53 & 12.12 \\
\midrule
\multirow{4}{*}{\shortstack[l]{\textbf{PACO-Part}\\1-shot (449/500)}} & \textbf{SegPFN} & 39.04 & 45.65 & 45.48 & 45.42 & 45.52 & 33.59 & 17.70 & 16.46 & 15.86 & 15.68 \\
 & FoRIS & 38.61 & 40.45 & 41.16 & 41.48 & 41.57 & 31.16 & 25.60 & 23.14 & 21.75 & 21.15 \\
 & FROST & 38.14 & 37.60 & 37.28 & 37.39 & 37.87 & 14.78 & 17.31 & 18.22 & 18.39 & 17.71 \\
 & INSID3 & 34.90 & 38.44 & 38.71 & 38.74 & 38.76 & 37.30 & 24.00 & 23.26 & 22.98 & 22.78 \\
\midrule
\multirow{4}{*}{\shortstack[l]{\textbf{PACO-Part}\\5-shot (445/500)}} & \textbf{SegPFN} & 46.17 & 53.86 & 55.41 & 56.19 & 56.54 & 31.89 & 15.63 & 13.78 & 12.68 & 12.23 \\
 & FoRIS & 44.42 & 46.53 & 48.36 & 48.94 & 49.19 & 34.34 & 28.80 & 24.79 & 22.67 & 21.67 \\
 & FROST & 51.13 & 51.68 & 51.71 & 51.69 & 51.89 & 10.44 & 10.95 & 11.18 & 11.22 & 10.82 \\
 & INSID3 & 40.76 & 46.53 & 46.46 & 46.42 & 46.33 & 37.92 & 19.69 & 19.30 & 19.08 & 18.97 \\
\bottomrule
\end{tabular*}
\end{table*}

For each benchmark, we assessed the stability of the difference in pooled foreground IoU between SegPFN and the highest-scoring published ICS baseline in Table~\ref{tab:personalized}.

We used groups of tasks as the resampling units for both the paired bootstrap and the leave-one-group-out analysis. On DAVIS-2017-ICS, each of the 20 videos formed one group. On CUHK-SYSU-ICS, we grouped tasks linked directly or indirectly by a shared image in either the reference or target role, yielding 109 groups. In each of 50{,}000 bootstrap draws, we sampled groups with replacement and recomputed each method's pooled foreground IoU on the same sample (Sec.~\ref{sec:supp-cohorts}). We used the 2.5th and 97.5th percentiles of the paired differences (SegPFN minus baseline) to obtain a 95\% confidence interval (CI). For the leave-one-group-out analysis, we omitted each group in turn and recomputed the difference.

On DAVIS-2017-ICS, SegPFN led FROST by $5.74$ percentage points (95\% CI, $0.60$ to $11.12$) and still led by at least $4.29$ points after omitting any one video. On CUHK-SYSU-ICS, the difference from Matcher was $-0.31$ points (95\% CI, $-6.16$ to $5.55$), and either method could lead depending on which group was omitted (Table~\ref{tab:supp-personalized-stability}).

\paragraph{Benchmark availability.}
We will release the task definitions for DAVIS-2017-ICS and CUHK-SYSU-ICS at \url{https://github.com/nyu-shenlab/SegPFN}. Each task identifies the reference and target images and their corresponding instance masks in the source dataset. No source images, videos, or mask files will be redistributed.

\section{Negative Reference Patches and Target Discrimination}
\label{sec:supp-negatives}
In the main paper, replacing hard-negative patches in SegPFN's context with ordinary background patches lowers mIoU and raises the non-target positive rate (NTPR) in all five settings (Table~\ref{tab:reference-negatives}).
Using these predictions as the 0\% and 100\% endpoints, we ask how much SegPFN recovers at intermediate retention levels and whether three training-free baselines respond similarly.

\paragraph{Retention protocol.}
We use the same 2,053 tasks and reference annotations for all four methods.
Eligible background patches overlap another instance of the target category in one-shot DAVIS-2017-ICS or another part of the same parent object in one- and five-shot PASCAL-Part and PACO-Part (Fig.~\ref{fig:reference-negatives}).
Each method retains 0, 10, 25, 50, or 100\% of its own eligible hard-negative patches, so absolute patch counts can differ. For INSID3 on DAVIS-2017-ICS, the 100\% endpoint uses the standard mIoU and DPR reported in Tables~\ref{tab:personalized} and~\ref{tab:false-positives}, respectively. DPR and NTPR are equivalent on this dataset.
For each PFN estimator, we replace omitted hard negatives with ordinary background patches, keeping both class counts and refinement settings fixed.
We exclude the omitted background tokens when constructing contrastive prototypes in FoRIS~\cite{hu2026foris}, estimating covariance and density in FROST~\cite{park2026frost}, or performing reciprocal matching in INSID3~\cite{cuttano2026insid3}.
We compute foreground mIoU on all tasks (Sec.~\ref{sec:supp-cohorts}). NTPR is the percentage of annotated non-target pixels predicted as foreground, averaged over tasks containing such pixels.

\begin{figure}[!htbp]
\centering
\begingroup
\definecolor{negseg}{HTML}{A13F77}
\definecolor{negforis}{HTML}{0072B2}
\definecolor{negfrost}{HTML}{C64E09}
\definecolor{neginsid}{HTML}{008171}
\pgfplotstableread{
retention SegPFN FoRIS FROST INSID3
0 33.24047634050667 31.119579336653842 50.34045976022775 30.18254708357392
10 56.423529123683394 32.29764719607497 51.09727638897768 39.60296315144802
25 57.45385641747897 35.07199233250309 50.879055344540845 40.024074709935
50 57.17686916540241 38.26064247917403 50.68275971755535 39.95958536945366
100 55.91852922170668 38.682211435845204 50.183633457942236 38.749141693115234
}\negmioudata
\pgfplotstableread{
retention SegPFN FoRIS FROST INSID3
0 66.43751450889218 73.47722874554755 10.473176820421717 69.71053666567526
10 24.26659789926238 71.03821169965397 12.318412577895899 45.27207348863721
25 16.672559912708465 66.27155808711231 13.879233792057786 43.07112447673133
50 14.746173074963197 60.172470994255875 15.246172183834236 40.71533259262039
100 13.56231464322763 55.196589013468426 17.127315110241963 37.72235677892413
}\negntprdata
\pgfplotsset{
  neg retention axis/.style={
    width=\dimexpr\columnwidth-28pt\relax,height=64pt,scale only axis,
    xmin=-2,xmax=102,xtick={0,10,25,50,100},
    axis lines*=left,axis line style={draw=black!55,line width=.4pt},
    tick align=outside,major tick length=2pt,tick style={draw=black!55},
    tick label style={font=\footnotesize},label style={font=\footnotesize},
    ymajorgrids,grid style={draw=black!14,line width=.3pt},
    title style={font=\small\bfseries,at={(0,1)},anchor=south west,inner sep=0pt,yshift=3pt},
  },
  neg SegPFN/.style={color=negseg,line width=1.25pt,mark=*,mark size=2pt},
  neg FoRIS/.style={color=negforis,line width=.9pt,mark=square,mark size=2.1pt},
  neg FROST/.style={color=negfrost,line width=.9pt,mark=triangle*,mark size=2.3pt},
  neg INSID/.style={color=neginsid,line width=.9pt,mark=diamond*,mark size=2.2pt},
}
\newcommand{\negretentiontraces}[1]{%
  \addplot[neg SegPFN] table[x=retention,y=SegPFN]{#1};
  \addplot[neg FoRIS] table[x=retention,y=FoRIS]{#1};
  \addplot[neg FROST] table[x=retention,y=FROST]{#1};
  \addplot[neg INSID] table[x=retention,y=INSID3]{#1};
}
\begin{tikzpicture}
\begin{axis}[neg retention axis,name=negmiou,title={(a) Foreground mIoU (\%) $\uparrow$},
  ymin=27,ymax=61,ytick={30,40,50,60},xticklabel=\empty]
  \negretentiontraces{\negmioudata}
\end{axis}
\begin{axis}[neg retention axis,at={(negmiou.south west)},anchor=north west,yshift=-22pt,
  ymin=0,ymax=80,ytick={0,20,40,60,80},title={(b) Non-target positive rate (\%) $\downarrow$},
  xlabel={Hard negatives retained (\%)},
  legend columns=4,legend cell align=left,
  legend style={at={(.5,0)},anchor=north,yshift=-30pt,font=\footnotesize,draw=none,
    inner sep=0pt,/tikz/every even column/.append style={column sep=3pt}},
  legend image post style={scale=.75}]
  \negretentiontraces{\negntprdata}
  \legend{SegPFN,FoRIS,FROST,INSID3}
\end{axis}
\end{tikzpicture}
\endgroup
\caption{\textbf{Hard-negative retention on DAVIS-2017-ICS.} We compare SegPFN, FoRIS, FROST, and INSID3 as the fraction of retained hard-negative reference patches increases from 0\% to 100\%. The top panel (a) shows foreground mIoU and the bottom panel (b) shows the non-target positive rate (NTPR).}
\label{fig:supp-hard-negative-retention}
\end{figure}
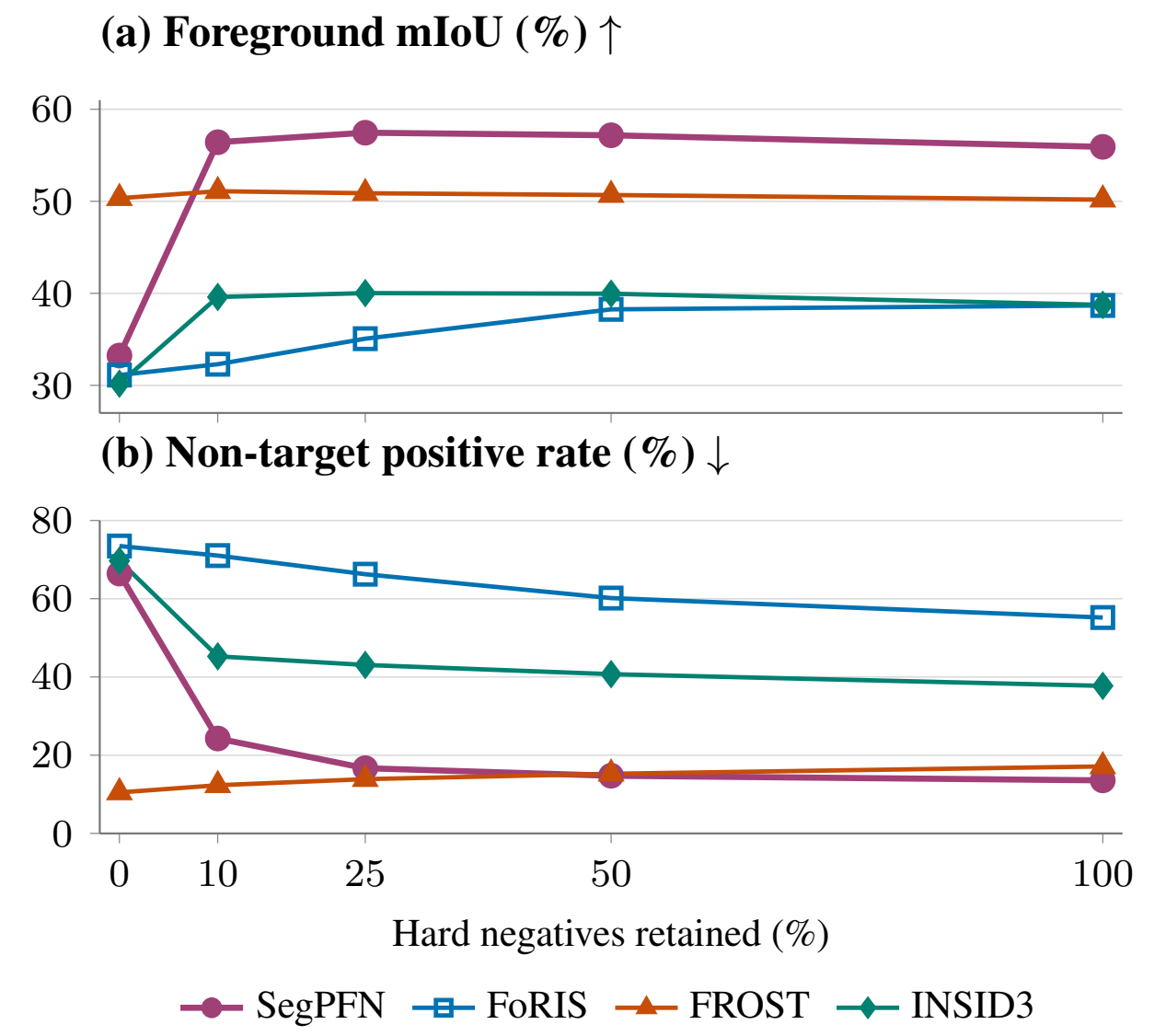

\filbreak
\paragraph{Effect of retention.}
Measured from each method's own 0\% score, SegPFN's mIoU gain at just 10\% retention exceeds every baseline's gain at full retention in all five settings (Table~\ref{tab:supp-hard-negative-retention}).
On DAVIS-2017-ICS, for example, restoring 10\% of SegPFN's sampled hard negatives raises mIoU by 23.18 points, compared with a maximum gain of 8.57 points for INSID3 among the three baselines at full retention (Fig.~\ref{fig:supp-hard-negative-retention}).
In absolute mIoU, SegPFN surpasses the strongest full-retention baseline in four of the five settings at 10\% retention and in all five at 25\%.
SegPFN also shows the largest NTPR reduction in four settings at 10\% retention and in all five at full retention.
FoRIS improves steadily, and INSID3 gains most of its mIoU by 10\% retention.
FROST already leads in mIoU in four settings when hard negatives are excluded. Restoring its full pool changes mIoU by at most 1.48 points and increases NTPR in four of the five settings, indicating that FROST derives little benefit from the additional hard-negative information.

SegPFN uses its available hard-negative reference patches more efficiently than any of the three baselines.
Retaining just 10\% of these patches recovers most of its mIoU gain, and each subsequent increase in retention further reduces false positives on related non-target instances or parts.

\section{Efficient Use of Annotated References}
\label{sec:supp-references}

FoRIS achieves the highest one-shot mIoU on semantic segmentation, while SegPFN leads both semantic and part segmentation in the five-shot setting (Tables~\ref{tab:semantic-results} and~\ref{tab:part-results}). We ask when the semantic ranking reverses and whether SegPFN can outperform five-shot baselines with fewer annotated references.\looseness=-1

\paragraph{Evaluation protocol.}
We evaluate SegPFN, FoRIS, FROST, and INSID3 on six semantic and two part segmentation datasets. Each $K$-shot episode pairs a target image with $K\in\{1,2,3,5\}$ annotated reference images. The two-, three-, and five-shot evaluations use the same targets and nested reference sets across all four methods. SegPFN uses the 133-dimensional input described in Sec.~\ref{sec:supp-segpfn-inference-settings} at all three shot counts. For one-shot evaluation, we use the original episodes and inference settings.

\begin{figure}[!htbp]
\centering
\begingroup
\definecolor{shotseg}{HTML}{A13F77}
\definecolor{shotforis}{HTML}{0072B2}
\definecolor{shotfrost}{HTML}{C64E09}
\definecolor{shotinsid}{HTML}{008171}
\pgfplotstableread{
K SegPFN FoRIS FROST INSID3
1 60.12213406032986 61.196928352779814 55.22210385004679 56.677129681905114
2 64.01184793034808 64.35120745527968 59.150736124933076 58.038896786802724
3 66.72855283558596 66.01619471069162 62.40183616613315 61.46953487183736
5 69.22 67.17664142184788 64.98064405653211 63.051667806837294
}\refsemanticdata
\pgfplotstableread{
K SegPFN FoRIS FROST INSID3
1 51.97399663925171 49.16512060165405 46.42050361633301 44.044188022613525
2 57.536152418692225 53.842216221075475 52.142353866751186 49.21519331181608
3 60.08743518873347 55.64490630846315 55.82714173191361 50.81561649457227
5 62.321916580200195 57.19996166229248 59.21898174285889 52.613457679748535
}\refpartdata
\pgfplotsset{
  ref count axis/.style={
    width=\dimexpr\columnwidth-35pt\relax, height=70pt, scale only axis,
    xmin=0.85, xmax=5.15, xtick={1,2,3,5},
    axis lines*=left, axis line style={draw=black!55,line width=.4pt},
    tick align=outside, major tick length=2pt, tick style={draw=black!55},
    tick label style={font=\footnotesize}, label style={font=\footnotesize},
    ylabel={mIoU (\%)}, ylabel style={yshift=-1pt},
    ymajorgrids, grid style={draw=black!14,line width=.3pt},
    title style={font=\small\bfseries,at={(0,1)},anchor=south west,inner sep=0pt,yshift=3pt},
  },
  ref SegPFN/.style={color=shotseg,line width=1.25pt,mark=*,mark size=2.0pt},
  ref FoRIS/.style={color=shotforis,line width=.9pt,mark=square,mark size=2.1pt},
  ref FROST/.style={color=shotfrost,line width=.9pt,mark=triangle*,mark size=2.3pt},
  ref INSID/.style={color=shotinsid,line width=.9pt,mark=diamond*,mark size=2.2pt},
}
\newcommand{\refcounttraces}[2]{%
  \pgfplotstablegetelem{3}{#2}\of{#1}%
  \edef\refcountbaseline{\pgfplotsretval}%
  \addplot[forget plot,no marks,black!55,densely dotted,line width=.65pt]
    coordinates {(0.85,\refcountbaseline) (5.15,\refcountbaseline)};
  \addplot[ref SegPFN] table[x=K,y=SegPFN]{#1};
  \addplot[ref FoRIS] table[x=K,y=FoRIS]{#1};
  \addplot[ref FROST] table[x=K,y=FROST]{#1};
  \addplot[ref INSID] table[x=K,y=INSID3]{#1};
}
\begin{tikzpicture}
\begin{axis}[ref count axis,name=refsem,title={(a) Semantic segmentation},
  ymin=52,ymax=74,ytick={55,60,65,70},xticklabel=\empty]
  \refcounttraces{\refsemanticdata}{FoRIS}
\end{axis}
\begin{axis}[ref count axis,at={(refsem.south west)},anchor=north west,yshift=-22pt,
  ymin=42,ymax=64,ytick={45,50,55,60},
  title={(b) Part segmentation},xlabel={Number of reference images ($K$)},
  legend columns=4,legend cell align=left,
  legend style={at={(.5,0)},anchor=north,yshift=-30pt,font=\footnotesize,draw=none,
    inner sep=0pt,/tikz/every even column/.append style={column sep=3pt}},
  legend image post style={scale=.75}]
  \refcounttraces{\refpartdata}{FROST}
  \legend{SegPFN,FoRIS,FROST,INSID3}
\end{axis}
\end{tikzpicture}
\endgroup
\caption{\textbf{Mean foreground mIoU by reference count.} Dotted lines mark the best five-shot baselines, FoRIS on semantic segmentation (67.18\%) and FROST on part segmentation (59.22\%).}
\label{fig:supp-reference-count-curves}
\end{figure}
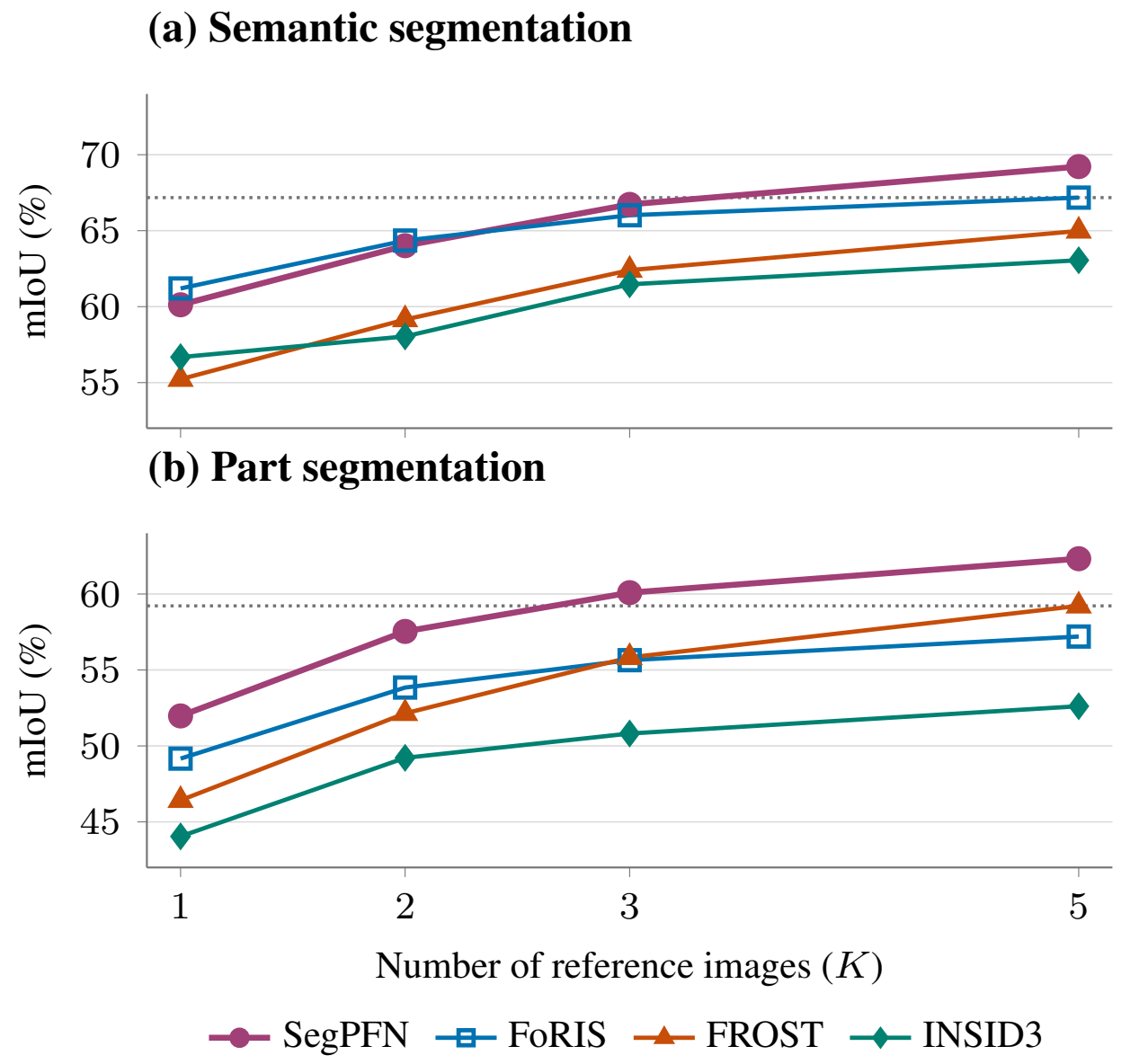

\begin{table}[!htbp]
\centering
\caption{\textbf{Foreground mIoU (\%) by number of annotated references.} Scores give equal weight to each dataset within a task. The one- and five-shot results are from Tables~\ref{tab:semantic-results} and~\ref{tab:part-results} of the main paper. Bold marks the highest mIoU in each row.}
\label{tab:supp-reference-count-scores}
\small
\setlength{\tabcolsep}{2.5pt}
\renewcommand{\arraystretch}{1.10}
\begin{tabular*}{\columnwidth}{@{\extracolsep{\fill}}lrrrr@{}}
\toprule
References & SegPFN & FoRIS & FROST & INSID3 \\
\midrule
\multicolumn{5}{@{}l}{\textit{Semantic segmentation (6 datasets)}} \\
1 & 60.12 & \textbf{61.20} & 55.22 & 56.68 \\
2 & 64.01 & \textbf{64.35} & 59.15 & 58.04 \\
3 & \textbf{66.73} & 66.02 & 62.40 & 61.47 \\
5 & \textbf{69.22} & 67.18 & 64.98 & 63.05 \\
\midrule
\multicolumn{5}{@{}l}{\textit{Part segmentation (2 datasets)}} \\
1 & \textbf{51.97} & 49.17 & 46.42 & 44.04 \\
2 & \textbf{57.54} & 53.84 & 52.14 & 49.22 \\
3 & \textbf{60.09} & 55.64 & 55.83 & 50.82 \\
5 & \textbf{62.32} & 57.20 & 59.22 & 52.61 \\
\bottomrule
\end{tabular*}
\end{table}

\begin{table*}[tp]
\centering
\begin{minipage}[t]{\dimexpr(\textwidth-\columnsep)/2\relax}
\vspace{0pt}
\centering
\caption{\textbf{Hyperparameter sensitivity (Overall mIoU, \%).} Both variants use cluster smoothing. The right pair adds reference-guided correction. Pre and Post denote scores before and after CRF refinement. A dagger ($\dagger$) marks the scikit-learn defaults for the listed hyperparameters given the chosen classifier and kernel. The bold value in the final column is the Overall mIoU reported for that family in Table~\ref{tab:inference-models}.}
\label{tab:supp-classifier-grid}
\small
\setlength{\tabcolsep}{1.5pt}
\renewcommand{\arraystretch}{1.02}
\begin{tabular*}{\linewidth}{@{\extracolsep{\fill}}lcrrrr@{}}
\toprule
 & & \multicolumn{2}{c}{Smoothing} & \multicolumn{2}{c}{+ Correction} \\
\cmidrule(lr){3-4}\cmidrule(l){5-6}
Predictor & Setting & Pre & Post & Pre & Post \\
\midrule
\multirow{2}{*}{\shortstack[l]{kNN\\$(k)$}} & $5^{\dagger}$ & 49.10 & 49.58 & 53.10 & \textbf{53.72} \\
 & $15$ & 46.49 & 46.86 & 52.60 & 53.22 \\
\midrule
\multirow{9}{*}{\shortstack[l]{RBF ridge\\$(\alpha,d\gamma)$}} & $(0.1, 0.1)$ & 51.85 & 52.39 & 52.86 & \textbf{53.42} \\
 & $(0.1, 1)$ & 34.87 & 35.16 & 42.57 & 43.03 \\
 & $(0.1, 10)$ & 11.01 & 10.95 & 16.57 & 16.60 \\
 & $(1, 0.1)$ & 49.64 & 50.19 & 52.14 & 52.72 \\
 & $(1, 1)^{\dagger}$ & 32.30 & 32.54 & 40.60 & 41.03 \\
 & $(1, 10)$ & 10.72 & 10.66 & 16.30 & 16.34 \\
 & $(10, 0.1)$ & 38.35 & 38.80 & 47.34 & 47.98 \\
 & $(10, 1)$ & 23.00 & 23.19 & 32.94 & 33.30 \\
 & $(10, 10)$ & 10.01 & 9.95 & 15.50 & 15.53 \\
\midrule
\multirow{3}{*}{\shortstack[l]{Linear ridge\\$(\alpha)$}} & $0.1$ & 51.21 & 51.84 & 52.48 & 53.10 \\
 & $1^{\dagger}$ & 51.26 & 51.90 & 52.51 & 53.13 \\
 & $10$ & 51.36 & 51.99 & 52.56 & \textbf{53.17} \\
\midrule
\multirow{3}{*}{\shortstack[l]{Logistic\\$(C)$}} & $0.1$ & 54.54 & 54.80 & 57.87 & \textbf{58.24} \\
 & $1^{\dagger}$ & 53.60 & 53.86 & 56.56 & 56.97 \\
 & $10$ & 53.24 & 53.63 & 55.85 & 56.38 \\
\midrule
\multicolumn{2}{@{}l}{TabPFN 3.5 (SegPFN)} & 59.06 & 59.48 & 60.13 & 60.67 \\
\bottomrule
\end{tabular*}
\end{minipage}\hfill%
\begin{minipage}[t]{\dimexpr(\textwidth-\columnsep)/2\relax}
\vspace{0pt}
\centering
\caption{\textbf{Foreground mIoU (\%) by dataset and reference count \(K\).} Each conventional predictor uses the same setting in every row, as selected in Table~\ref{tab:supp-classifier-grid}. All predictors use smoothing, reference-guided correction, and CRF. Bold marks the highest score in each row.}
\label{tab:supp-classifier-datasets}
\small
\setlength{\tabcolsep}{1.35pt}
\renewcommand{\arraystretch}{1.02}
\begin{tabular*}{\linewidth}{@{\extracolsep{\fill}}lcrrrrr@{}}
\toprule
Dataset & $K$ & kNN & \shortstack{RBF\\ridge} & \shortstack{Linear\\ridge} & Logistic & SegPFN \\
\midrule
\multicolumn{7}{@{}l}{\emph{Semantic}} \\
\multirow{2}{*}{COCO-$20^i$} & 1 & 44.49 & 52.88 & 52.28 & 53.95 & \textbf{57.62} \\
 & 5 & 61.03 & 68.46 & 67.95 & 69.00 & \textbf{69.51} \\
\multirow{2}{*}{LVIS-$92^i$} & 1 & 37.13 & 39.96 & 40.05 & 40.42 & \textbf{46.32} \\
 & 5 & 47.81 & 47.28 & 46.15 & \textbf{54.03} & 53.84 \\
\multirow{2}{*}{ISIC} & 1 & 51.74 & 59.35 & 59.29 & 54.93 & \textbf{60.46} \\
 & 5 & 61.66 & 65.31 & 65.39 & 65.74 & \textbf{67.93} \\
\multirow{2}{*}{Lung} & 1 & 85.15 & 84.60 & 84.25 & 86.08 & \textbf{86.53} \\
 & 5 & 85.30 & 84.97 & 84.70 & 86.61 & \textbf{86.92} \\
\multirow{2}{*}{iSAID} & 1 & 45.62 & 51.74 & 50.51 & 52.26 & \textbf{56.10} \\
 & 5 & 50.48 & 60.83 & 60.36 & 62.40 & \textbf{65.41} \\
\multirow{2}{*}{SUIM} & 1 & 41.08 & 58.76 & 59.34 & 51.96 & \textbf{59.61} \\
 & 5 & 55.40 & 69.42 & 69.39 & 68.24 & \textbf{70.14} \\
\midrule
\multicolumn{7}{@{}l}{\emph{Part}} \\
\multirow{2}{*}{PASCAL-Part} & 1 & 49.27 & 52.88 & 52.19 & 52.46 & \textbf{58.28} \\
 & 5 & 60.64 & 61.01 & 60.96 & 66.16 & \textbf{67.19} \\
\multirow{2}{*}{PACO-Part} & 1 & 37.16 & 38.19 & 38.00 & 41.49 & \textbf{47.34} \\
 & 5 & 49.17 & 46.98 & 46.61 & 56.85 & \textbf{58.55} \\
\midrule
\multicolumn{7}{@{}l}{\emph{Personalized}} \\
PerMIS & 1 & 65.64 & 57.15 & 59.27 & \textbf{73.57} & 71.76 \\
DAVIS-2017-ICS & 1 & 51.54 & 40.54 & 39.33 & 52.72 & \textbf{55.92} \\
CUHK-SYSU-ICS & 1 & \textbf{52.43} & 47.89 & 46.72 & 48.74 & 49.71 \\
\bottomrule
\end{tabular*}
\end{minipage}
\end{table*}

\filbreak
\paragraph{Performance across shot counts.}
On semantic segmentation, SegPFN trails FoRIS by 0.34 mIoU points in the two-shot setting. The ranking reverses in the three-shot comparison (66.73\% for SegPFN vs.\ 66.02\% for FoRIS), and SegPFN's five-shot lead reaches 2.04 points (Fig.~\ref{fig:supp-reference-count-curves} and Table~\ref{tab:supp-reference-count-scores}). On part segmentation, SegPFN leads at all four shot counts. Its three-shot mIoU (60.09\%) already exceeds the best five-shot baseline (FROST, 59.22\%).

Three-shot SegPFN outperforms all three five-shot baselines on part segmentation and two of the three on semantic segmentation. Five-shot FoRIS retains a 0.45-point lead on semantic segmentation (67.18\% vs.\ 66.73\%).\par

\section{Comparison with Conventional Classifiers}
\label{sec:supp-classifiers}

\begin{table*}[tp]
\centering
\caption{\textbf{Dataset-level prediction-refinement ablation on DINOv3-L under the main-benchmark protocol.} Values are foreground mIoU (\%). Within each shot setting, the columns add cluster smoothing and then reference-guided correction to the PFN baseline. Bold marks the highest score in each row for that shot setting. Parentheses give personalized task counts, and dashes mark unevaluated settings.}
\label{tab:supp-component-ablation}
\label{tab:supp-personalized-pfn}
\small
\setlength{\tabcolsep}{4pt}
\renewcommand{\arraystretch}{1.10}
\begin{tabular*}{\textwidth}{@{\extracolsep{\fill}}lcccccc@{}}
\toprule
\multirow{2}{*}{Dataset} & \multicolumn{3}{c}{1-shot} & \multicolumn{3}{c}{5-shot} \\
\cmidrule(lr){2-4}\cmidrule(l){5-7}
& \shortstack{PFN\\baseline} & \shortstack{+ Cluster\\smoothing} & \shortstack{+ Reference-guided\\correction} & \shortstack{PFN\\baseline} & \shortstack{+ Cluster\\smoothing} & \shortstack{+ Reference-guided\\correction} \\
\midrule
\multicolumn{7}{@{}l}{\textit{Semantic segmentation}} \\
COCO-$20^i$ & 51.17 & 54.83 & \textbf{58.35} & 66.71 & 68.35 & \textbf{68.63} \\
LVIS-$92^i$ & 40.09 & 41.38 & \textbf{42.57} & 56.72 & \textbf{57.08} & 56.67 \\
ISIC & 52.04 & 56.67 & \textbf{60.46} & 63.24 & 65.64 & \textbf{67.93} \\
Lung & \textbf{89.04} & 87.37 & 86.53 & \textbf{90.35} & 88.01 & 86.92 \\
iSAID & 47.89 & 50.46 & \textbf{53.47} & 63.27 & 64.74 & \textbf{65.48} \\
SUIM & 53.61 & 55.78 & \textbf{59.36} & 66.20 & 67.91 & \textbf{69.66} \\
\addlinespace[3pt]
\multicolumn{7}{@{}l}{\textit{Part segmentation}} \\
PASCAL-Part & 52.04 & 55.57 & \textbf{57.45} & 66.24 & \textbf{67.43} & 67.09 \\
PACO-Part & 39.79 & 44.19 & \textbf{46.50} & 54.64 & 57.34 & \textbf{57.55} \\
\addlinespace[3pt]
\multicolumn{7}{@{}l}{\textit{Personalized segmentation}} \\
PerMIS (432) & \textbf{74.19} & 72.33 & 71.76 & \textemdash & \textemdash & \textemdash \\
DAVIS-2017-ICS (53) & 54.20 & 53.51 & \textbf{55.92} & \textemdash & \textemdash & \textemdash \\
CUHK-SYSU-ICS (151) & 48.63 & 48.70 & \textbf{49.71} & \textemdash & \textemdash & \textemdash \\
\bottomrule
\end{tabular*}
\end{table*}

\paragraph{Comparison protocol.}
To supplement the PFN inference ablation in Table~\ref{tab:inference-models}, we report all 17 configurations tested for distance-weighted kNN, linear ridge, RBF ridge, and logistic regression (Table~\ref{tab:supp-classifier-grid}). Across the $15{,}036$ episodes in Table~\ref{tab:supp-backbone-cohort}, all predictors use the same frozen DINOv3-L features, reference-fitted PCA projections, sampled reference contexts, and mask refinement. For each conventional classifier, we standardize features using only the sampled reference patches. Both ridge predictors fit labels in $\{-1,+1\}$ and convert scores to foreground probabilities with a fixed sigmoid. For RBF ridge, $d$ in $d\gamma$ denotes the input dimension. Within each family, we select the configuration with the highest Overall mIoU after cluster smoothing, reference-guided correction, and CRF. Selection and evaluation use the same episodes, and each selected configuration applies to all datasets and shot settings.

\paragraph{Results.}
Lowering logistic regression's $C$ from $1$ to $0.1$ raises its Overall mIoU from 56.97 to 58.24, the highest score among the conventional classifiers. SegPFN reaches 60.67, a 2.43-point advantage, and leads all five task--shot averages. It also achieves the highest mIoU in 16 of the 19 dataset--shot comparisons (Table~\ref{tab:supp-classifier-datasets}).

\section{Dataset-Level Effects of Prediction Refinement}
\label{sec:supp-component-ablation}
\paragraph{Comparison protocol.}
To supplement the prediction refinement ablation in Table~\ref{tab:ablations}, we report mIoU by dataset and shot setting under the same DINOv3-L evaluation protocol. Starting from the PFN baseline, we add cluster smoothing and then reference-guided correction. All three configurations retain min--max normalization and CRF.

\paragraph{Results.}
Full refinement yields larger gains in one-shot than in five-shot evaluation for both semantic and part segmentation (Table~\ref{tab:supp-component-ablation}). Across the six semantic datasets, the mean gain over the PFN baseline is 4.48 mIoU points with one reference and 1.47 points with five. Across the two part datasets, the corresponding gains are 6.06 and 1.88 points. Lung favors the PFN baseline at both shot counts. Overall, full refinement improves mIoU in 15 of 19 dataset--shot comparisons, including two of the three one-shot personalized benchmarks (DAVIS-2017-ICS and CUHK-SYSU-ICS).

\section{Robustness of SegPFN Across Visual Backbone Sizes}
\label{sec:supp-backbone}
\begin{table*}[tp]
\centering
\caption{\textbf{Foreground mIoU (\%) across DINOv3 backbone sizes on the matched evaluation subsets.} Parentheses in the task headings give dataset counts. The indented SegPFN rows disable reference-guided correction alone or both correction and cluster smoothing. All three SegPFN configurations retain min--max normalization and CRF. Overall weights the three tasks equally after averaging shot settings within each task. Bold and underline mark the best and second-best scores within each backbone and column.}
\label{tab:supp-backbone-results}
\small
\setlength{\tabcolsep}{4pt}
\renewcommand{\arraystretch}{1.02}
\begin{tabular*}{\textwidth}{@{\extracolsep{\fill}}llrrrrrr@{}}
\toprule
\multirow{2}{*}{Backbone} & \multirow{2}{*}{Method} & \multicolumn{2}{c}{Semantic (6)} & \multicolumn{2}{c}{Part (2)} & Personalized (3) & \multirow{2}{*}{Overall} \\
\cmidrule(lr){3-4}\cmidrule(lr){5-6}\cmidrule(lr){7-7}
& & 1-shot & 5-shot & 1-shot & 5-shot & 1-shot & \\
\midrule
\multirow{6}{*}{DINOv3-S} & INSID3~\cite{cuttano2026insid3} & 32.38 & 36.27 & 34.01 & 39.70 & 26.78 & 32.65 \\
 & FROST~\cite{park2026frost} & 46.70 & 56.57 & 35.22 & 45.21 & \textbf{46.74} & 46.20 \\
 & FoRIS~\cite{hu2026foris} & \underline{48.81} & 58.35 & 37.57 & 44.08 & 35.18 & 43.20 \\
 & SegPFN (default) & 48.55 & 60.26 & \underline{39.61} & 47.81 & 39.93 & 46.01 \\
 & \hspace{0.7em}\textit{w/o reference-guided correction} & \textbf{49.72} & \textbf{62.13} & \textbf{40.90} & \textbf{53.34} & 42.37 & \underline{48.47} \\
 & \hspace{0.7em}\textit{w/o correction and smoothing} & 48.42 & \underline{61.80} & 39.17 & \underline{53.22} & \underline{45.71} & \textbf{49.00} \\
\midrule
\multirow{6}{*}{DINOv3-B} & INSID3~\cite{cuttano2026insid3} & 39.70 & 45.24 & 35.10 & 43.75 & 39.50 & 40.47 \\
 & FROST~\cite{park2026frost} & 51.76 & 61.84 & 42.47 & 55.52 & \underline{53.11} & 52.97 \\
 & FoRIS~\cite{hu2026foris} & 52.38 & 59.63 & 42.05 & 49.77 & 42.34 & 48.08 \\
 & SegPFN (default) & \underline{55.05} & 64.48 & \underline{45.44} & 56.89 & 51.67 & 54.20 \\
 & \hspace{0.7em}\textit{w/o reference-guided correction} & \textbf{55.22} & \textbf{66.44} & \textbf{46.31} & \textbf{59.93} & 52.46 & \textbf{55.47} \\
 & \hspace{0.7em}\textit{w/o correction and smoothing} & 53.32 & \underline{65.65} & 43.55 & \underline{57.95} & \textbf{53.83} & \underline{54.69} \\
\midrule
\multirow{6}{*}{DINOv3-L} & INSID3~\cite{cuttano2026insid3} & 57.76 & 63.41 & 44.63 & 53.64 & 49.61 & 53.11 \\
 & FROST~\cite{park2026frost} & 55.10 & 64.74 & 45.91 & 59.49 & 53.93 & 55.52 \\
 & FoRIS~\cite{hu2026foris} & \textbf{62.04} & 67.89 & 49.17 & 57.58 & 49.43 & 55.92 \\
 & SegPFN (default) & \underline{61.11} & \textbf{68.96} & \textbf{52.81} & \underline{62.87} & \textbf{59.13} & \textbf{60.67} \\
 & \hspace{0.7em}\textit{w/o reference-guided correction} & 58.82 & \underline{68.40} & \underline{50.35} & \textbf{62.94} & 58.18 & \underline{59.48} \\
 & \hspace{0.7em}\textit{w/o correction and smoothing} & 56.62 & 67.38 & 46.02 & 60.86 & \underline{59.01} & 58.15 \\
\bottomrule
\end{tabular*}
\end{table*}

To test whether SegPFN retains its advantage with smaller visual backbones, we compare it with FoRIS, FROST, and INSID3 on DINOv3-S, DINOv3-B, and the DINOv3-L backbone used in the main experiments. All four methods use the same backbone at each size, and SegPFN uses the same frozen PFN throughout.

\paragraph{Comparison protocol.}
We use the subsets listed in Table~\ref{tab:supp-backbone-cohort} and score predictions as described in Sec.~\ref{sec:supp-cohorts}. Within each shot setting, we keep the target images, annotated references, and reference order fixed across methods and backbone sizes. We score masks at the native resolution of each target image on DAVIS-2017-ICS and CUHK-SYSU-ICS and on a $1024\times1024$ grid for all other datasets. At each backbone size, we also evaluate SegPFN with different refinement configurations (Sec.~\ref{sec:supp-component-ablation}).

\filbreak
\paragraph{Results.}
With its default configuration, SegPFN outperforms all three competing methods on five-shot semantic and part segmentation at every backbone size (Table~\ref{tab:supp-backbone-results}). In Overall mIoU, it trails FROST by 0.19 points on DINOv3-S (46.01 vs.\ 46.20). It leads the strongest competitor by 1.23 points on DINOv3-B (54.20 vs.\ 52.97 for FROST) and by 4.75 points on DINOv3-L (60.67 vs.\ 55.92 for FoRIS).


The ablation rows in Table~\ref{tab:supp-backbone-results} show that adding cluster smoothing to the PFN baseline improves semantic and part mIoU at both shot counts on all three backbones and lowers personalized mIoU at every size. Adding reference-guided correction after smoothing reduces Overall mIoU by 2.46 points on DINOv3-S and 1.27 on DINOv3-B, and raises it by 1.19 points on DINOv3-L.

At every backbone size, both SegPFN configurations without reference-guided correction outperform the strongest competing method in Overall mIoU. The PFN baseline leads by 1.72--2.80 points, and SegPFN with cluster smoothing as the only added refinement step leads by 2.27--3.56 points.


\section{Inference Efficiency}
\label{sec:supp-efficiency}
We compare inference time per evaluation task for SegPFN, FoRIS, FROST, and INSID3. In each task, the method predicts a mask for one target image using one or five annotated reference images.

\begin{samepage}
\paragraph{Timing protocol.}
For each shot setting, we use ten tasks from each of the six semantic and two part segmentation datasets. All four methods process the same 80 tasks one at a time on a single NVIDIA A100 PCIe GPU (80\,GB). After warm-up, timing starts when a method reads cached visual features from disk and ends when the final mask reaches the CPU. We average three runs per task and report the median across the 80 tasks.\par
\end{samepage}

\begin{samepage}
\paragraph{Results.}
SegPFN takes a median of 4.48\,s per task with one reference and 7.00\,s with five (Table~\ref{tab:supp-inference-latency}). Although SegPFN improves Overall mIoU by 4.70 points over the strongest baseline on the main benchmarks (Sec.~\ref{sec:generalization}), its current implementation takes longer than all three baselines in both settings.\par
\end{samepage}

\begin{table}[!htbp]
\centering
\caption{\textbf{Inference latency with cached visual features.} We report the median time per task in seconds. Bold marks the fastest method for each reference count.}
\label{tab:supp-inference-latency}
\small
\begin{tabular*}{\columnwidth}{@{\extracolsep{\fill}}lrr@{}}
\toprule
Method & 1-shot & 5-shot \\
\midrule
FoRIS & 0.99 & 1.98 \\
FROST & \textbf{0.47} & 2.72 \\
INSID3 & 0.61 & \textbf{1.18} \\
\midrule
SegPFN & 4.48 & 7.00 \\
\bottomrule
\end{tabular*}
\end{table}

\filbreak
\section{Conceptual Contribution and Implications}
\label{sec:supp-scope}

\textbf{SegPFN's conceptual contribution is to reformulate ICS as in-context tabular classification, enabling a pretrained tabular learner to infer what to segment from annotated references without segmentation-specific training.} A frozen visual encoder extracts patch features from the reference and target images. Using the reference features and their foreground/background labels as context, a frozen PFN predicts labels for the target patches, supporting semantic, part, and personalized segmentation within a single framework. Visual representations and in-context inference can thus be pretrained in different domains and brought together through a common classification formulation.

SegPFN uses reference information more efficiently than INSID3, FoRIS, and FROST, whose inference procedures rely on manually designed matching, aggregation, or statistical estimation~\cite{cuttano2026insid3,hu2026foris,park2026frost}. In part segmentation, SegPFN with three annotated references outperforms the strongest baseline using five (Sec.~\ref{sec:supp-references}). In all five dataset--shot settings, SegPFN retaining 25\% of its sampled hard negatives achieves higher mIoU than each baseline using its full hard-negative pool (Sec.~\ref{sec:supp-negatives}). Across 11 datasets, SegPFN improves Overall mIoU by 4.70 points over the strongest overall baseline (Sec.~\ref{sec:generalization}). By transferring in-context inference through task reformulation, SegPFN provides a simple and effective new paradigm for training-free ICS.

\end{document}